\documentclass{article}
\usepackage{microtype}
\usepackage{graphicx}
\usepackage{wrapfig}
\usepackage{booktabs} 
\usepackage{makecell}
\usepackage{graphicx}
\usepackage{caption}
\usepackage{courier}
\usepackage{multirow}
\usepackage{color}
\usepackage{adjustbox}
\usepackage{subcaption}
\usepackage{amssymb}
\usepackage{amsmath}
\usepackage{gensymb}
\usepackage{amsfonts}       
\usepackage{enumitem}
\usepackage{multirow}
\usepackage{algorithm}
\usepackage[noend]{algorithmic} 
\usepackage{amsmath,nccmath}
\usepackage{amsmath,amsfonts,amssymb,bbm, bm}
\usepackage[dvipsnames]{xcolor}
\usepackage{hyperref}

\usepackage{bbm}
\usepackage[accepted]{icml2022}

\icmltitlerunning{Improving Out-of-Distribution Robustness via Selective Augmentation}

\usepackage{hyperref}
\usepackage{url}

\newcommand{\yao}[1]{\textcolor{black}{#1}}

\def\gam{\gamma}
\def\Sig{\Sigma}
\def\eps{\epsilon}
\newtheorem{theorem}{Theorem}

\def\R{\mathbbm{R}}
\def\P{\mathbbm{P}}
\def\E{\mathbbm{E}}
\def\lam{\lambda}
\def\Sig{\Sigma}
\def\tx{\tilde{x}}
\def\ty{\tilde{y}}

\begin{document}
\twocolumn[
\icmltitle{Improving Out-of-Distribution Robustness via Selective Augmentation}

\icmlsetsymbol{equal}{*}

\begin{icmlauthorlist}
\icmlauthor{Huaxiu Yao}{equal,stanford}
\icmlauthor{Yu Wang}{equal,ucsd}
\icmlauthor{Sai Li}{ruc}
\icmlauthor{Linjun Zhang}{rutgers}
\icmlauthor{Weixin Liang}{stanford}
\\
\icmlauthor{James Zou}{stanford}
\icmlauthor{Chelsea Finn}{stanford}
\end{icmlauthorlist}

\icmlaffiliation{ucsd}{University of California San Diego, CA, USA}
\icmlaffiliation{ruc}{Renmin University of China, Beijing, China}
\icmlaffiliation{rutgers}{Rutgers University, NJ, USA}
\icmlaffiliation{stanford}{Stanford University, CA, USA}

\icmlcorrespondingauthor{Huaxiu Yao}{huaxiu@cs.stanford.edu}
\icmlcorrespondingauthor{Sai Li}{saili@ruc.edu.cn}

\icmlkeywords{Machine Learning, ICML}

\vskip 0.3in

]

\printAffiliationsAndNotice{\icmlEqualContribution. This work was done when Yu Wang was mentored by Huaxiu Yao remotely.}

\begin{abstract}
Machine learning algorithms typically assume that training and test examples are drawn from the same distribution. However, distribution shift is a common problem in real-world applications and can cause models to perform dramatically worse at test time. In this paper, we specifically consider the problems of subpopulation shifts (e.g., imbalanced data) and domain shifts. While prior works often seek to explicitly regularize internal representations or predictors of the model to be domain invariant, we instead aim to learn invariant predictors without restricting the model's internal representations or predictors. This leads to a simple mixup-based technique which learns invariant predictors via selective augmentation called LISA. LISA selectively interpolates samples either with the same labels but different domains or with the same domain but different labels. Empirically, we study the effectiveness of LISA on nine benchmarks ranging from subpopulation shifts to domain shifts, and we find that LISA consistently outperforms other state-of-the-art methods and leads to more invariant predictors. We further analyze a linear setting and theoretically show how LISA leads to a smaller worst-group error. Code is released in \href{https://github.com/huaxiuyao/LISA}{https://github.com/huaxiuyao/LISA}
\end{abstract}
\section{Introduction}
To deploy machine learning algorithms in real-world applications, we must pay attention to distribution shift, i.e. when the test distribution is different from the training distribution, which substantially degrades model performance. In this paper, we refer this problem as out-of-distribution (OOD) generalization and specifically consider performance gaps caused by two kinds of distribution shifts: \emph{subpopulation shifts} and \emph{domain shifts}. In subpopulation shifts, the test domains (or subpopulations) are seen but underrepresented in the training data. When subpopulation shift occurs, models may perform poorly when they falsely rely on spurious correlations between the particular subpopulation and the label. For example, in health risk prediction, a machine learning model trained on the entire population may associate the labels with demographic features (e.g., gender and age), making the model fail on the test set when such an association does not hold in reality. In domain shifts, the test data is from new domains, which requires the trained model to generalize well to test domains without seeing the data from those domains at training time. In the health risk example, we may want to train a model on patients from a few sampled hospitals and then deploy the model to a broader set of hospitals~\citep{koh2021wilds}. 

To improve model robustness under these two kinds of distribution shifts, prior works have proposed various regularizers to learn representations or predictors that are invariant to different domains while still containing sufficient information to fulfill the task~\citep{li2018domain,sun2016deep,arjovsky2019invariant,krueger2021out,rosenfeld2020risks}. However, designing regularizers that are widely suitable to datasets from diverse domains is challenging, and unsuitable regularizers may adversely limit the model's expressive power or yield a difficult optimization problem, leading to inconsistent performance among various real-world datasets. For example, on the WILDS datasets, invariant risk minimization (IRM)~\citep{arjovsky2019invariant} with reweighting -- a representative method for learning invariant predictor -- outperforms empirical risk minimization (ERM) on CivilComments, but fails to improve robustness on a variety of other datasets like Camelyon17 and RxRx1~\citep{koh2021wilds}. 

\begin{figure*}[t]
\centering
  \includegraphics[width=0.95\textwidth]{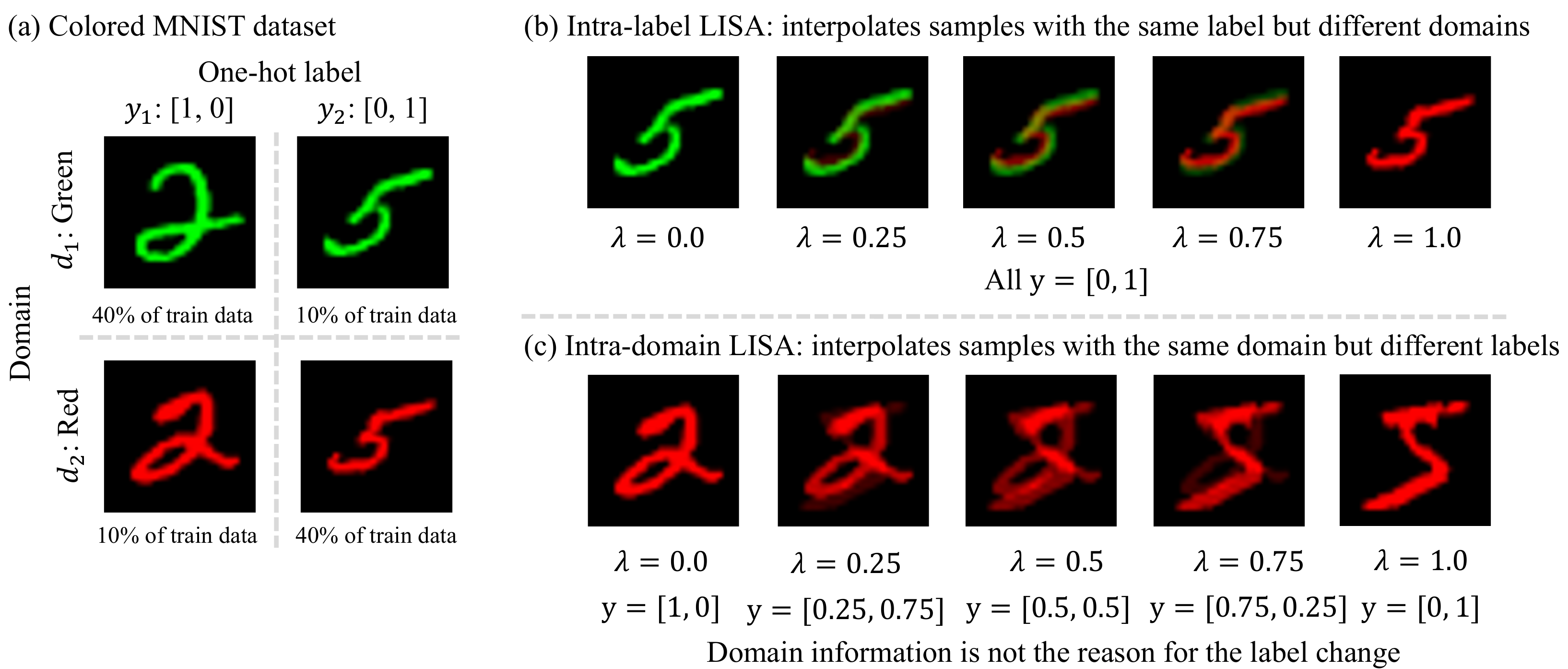}
  \vspace{-0.5em}
  \caption{Illustration of the variants of LISA (Intra-label LISA and Intra-domain LISA) on Colored MNIST dataset. $\lambda$ represents the interpolation ratio, which is sampled from a Beta distribution. (a) Colored MNIST (CMNIST). We classify MNIST digits as two classes, and original digits (0,1,2,3,4) and (5,6,7,8,9) are labeled as class 0 and 1, respectively. Digit color is used as domain information, which is spuriously correlated with labels in training data; (b) Intra-label LISA (LISA-L) cancels out spurious correlation by interpolating samples with the same label; (c) Intra-domain LISA (LISA-D) interpolates samples with the same domain but different labels to encourage the model to learn specific features within a domain.
  }\label{fig:method_illustration}
  \vspace{-1.5em}
\end{figure*}

\yao{Instead of explicitly imposing regularization, we propose to learn invariant predictors through data interpolation, leading to a simple algorithm called \textbf{LISA} (\textbf{L}earning \textbf{I}nvariant Predictors with \textbf{S}elective \textbf{A}ugmentation).} Concretely, inspired by mixup~\citep{zhang2017mixup}, LISA linearly interpolates the features for a pair of samples and applies the same interpolation strategy on the corresponding labels. Critically, the pairs are selectively chosen according to two selective augmentation strategies -- intra-label LISA (LISA-L) and intra-domain LISA (LISA-D), which are described below and illustrated on Colored MNIST dataset in Figure~\ref{fig:method_illustration}. Intra-label LISA (Figure~\ref{fig:method_illustration}(b)) interpolates samples with the same label but from different domains, aiming to eliminate domain-related spurious correlations. Intra-domain LISA (Figure~\ref{fig:method_illustration}(c)) interpolates samples with the same domain but different labels, such that the model should learn to ignore the domain information and generate different predicted values as the interpolation ratio changes. In this way, LISA encourages the model to learn domain-invariant predictors without any explicit constraints or regularizers. 

The \textbf{primary contributions} of this paper are as follows: (1) We propose a simple yet widely-applicable method for learning domain invariant predictors that is shown to be robust to subpopulation shifts and domain shifts. (2) We conduct broad experiments to evaluate LISA on nine benchmark datasets from diverse domains. In these experiments, we make the following key observations. First, we observe that LISA consistently outperforms seven prior methods to address subpopulation and domain shifts. Second, we find that LISA produces predictors that are consistently more domain invariant than prior approaches. Third, we identify that the performance gains of LISA are from canceling out domain-specific information or spurious correlations and learning invariant predictors, rather than simply involving more data via interpolation. Finally, when the degree of distribution shift increases, LISA achieves more significant performance gains. (3) We provide a theoretical analysis of the phenomena distilled from the empirical studies, where we provably demonstrate that LISA can mitigate spurious correlations and therefore lead to smaller worst-domain error compared with ERM and vanilla mixup. We also note that to the best of our knowledge, our work provides the first theoretical framework of studying how mixup (with or without the selective augmentation strategies) affects mis-classification error. 
\section{Preliminaries}
\label{sec:prelim}
In this paper, we consider the setting where one predicts the label $y \in \mathcal{Y}$ based on the input feature $x\in \mathcal{X}$. Given a parameter space $\Theta$ and a loss function $\ell$, we need to train a model $f_{\theta}$ under the training distribution $P_{tr}$, where $\theta \in \Theta$. In empirical risk minimization (ERM), the empirical distribution over the training data is $\hat{P}_{tr}$; ERM optimizes the following objective:
\begin{equation}
\label{eq:erm}
\theta^{*} := \arg\min_{\theta \in \Theta} \mathbb{E}_{(x,y)\sim \hat{P}} [\ell(f_{\theta}(x), y)].
\end{equation}
In a traditional machine learning setting, a test set, sampled from a test distribution $P_{ts}$, is used to evaluate the generalization of the trained model $\theta^{*}$, where the test distribution is assumed to be the same as the training distribution, i.e., $P^{tr}=P^{ts}$. In this paper, we are interested in the setting when distribution shift occurs, i.e., $P^{tr}\neq P^{ts}$.

Specifically, following~\citet{muandet2013domain,albuquerque2019generalizing,koh2021wilds}, we regard the overall data distribution containing $\mathcal{D}=\{1,\ldots,D\}$ domains and each domain $d\in \mathcal{D}$ is associated with a data distribution $P_d$ over a set $(X, Y, d)=\{(x_i, y_i, d)\}_{i=1}^{N^d}$, where $N^d$ is the number of samples in domain $d$. Then, we formulate the training distribution as the mixture of $D$ domains, i.e., $P^{tr}=\sum_{d\in \mathcal{D}}r_d^{tr} P_d$, where $\{r_d^{tr}\}$ denotes the mixture probabilities in training set. Here, the training domains are defined as $\mathcal{D}^{tr}=\{d\in \mathcal{D} | r_d^{tr}>0\}$. Similarly, the test distribution could be represented as $P^{ts}=\sum_{d\in \mathcal{D}}r_d^{ts} P_d$, where $\{r_d^{ts}\}$ is the mixture probabilities in test set. The test domains are defined as $\mathcal{D}^{ts}=\{d\in \mathcal{D} | r_d^{ts}>0\}$.

In subpopulation shifts, the test set has domains that have been seen in the training set, but with a different proportion of subpopulations, i.e., $\mathcal{D}^{ts} \subseteq \mathcal{D}^{tr}$ but $\{r^{ts}_d\}\neq \{r^{tr}_d\}$. Under this setting, following~\citet{sagawa2019distributionally}, we consider group-based spurious correlations, where each group $g \in \mathcal{G}$ is defined to be associated with a domain $d$ and a label $y$, i.e., $g=(d,y)$. We assume that the domain is spuriously correlated with the label. For example, we illustrate the CMNIST dataset in Figure~\ref{fig:method_illustration}, where the digit color $d$ (green or red) is spuriously correlated with the label $y$ ([1, 0] or [0, 1]). 
Based on the group definition, we evaluate the model via the worst test group error, i.e., $\max_{g}\mathbb{E}_{(x,y)\sim g}[\ell_{0-1}(f_{\theta}(x),y)]$, where $\ell_{0-1}$ represents the 0-1 loss.

In domain shifts, we investigate the problem where the test domains are disjoint from the training domains, i.e., $\mathcal{D}^{tr} \cap \mathcal{D}^{ts} = \emptyset$. In general, we assume the test domains share some common properties with the training domains. For example, in Camelyon17~\citep{koh2021wilds}, we train the model on some hospitals and test it in a new hospital. We evaluate the worst-domain and/or average performance of the classifier across all test domains.

\section{Learning Invariant Predictors with Selective Augmentation}
\label{sec:method}
This section presents LISA, a simple way to improve robustness to subpopulation shifts and domain shifts. The key idea behind LISA is to encourage the model to learn invariant predictors by selective data interpolation, which could also alleviates the effects of domain-related spurious correlations. Before detailing how to select interpolated samples, we first provide a general formulation for data interpolation.

In LISA, we perform linear interpolation between training samples. Specifically, given samples $(x_i, y_i, d_i)$ and $(x_j, y_j, d_j)$ drawn from domains $d_i$ and $d_j$, we apply mixup~\citep{zhang2017mixup}, a simple data interpolation strategy, separately on the input features and corresponding labels as:
\begin{equation}
\label{eq:mix}
    x_{mix}=\lambda x_i + (1-\lambda) x_j,\; y_{mix} = \lambda y_i + (1-\lambda) y_j,
\end{equation}
where the interpolation ratio $\lambda \in [0,1]$ is sampled from a Beta distribution $\rm{Beta}(\alpha, \beta)$ and $y_i$ and $y_j$ are one-hot vectors for classification problem. Notice that the mixup approach in~\eqref{eq:mix} can be replaced by CutMix~\citep{yun2019cutmix}, which shows stronger empirical performance in vision-based applications. In text-based applications, we can use Manifold Mixup~\citep{verma2019manifold}, interpolating the representations of a pre-trained model, e.g., the output of BERT~\citep{devlin2018bert}.

After obtaining the interpolated features and labels, we replace the original features and labels in ERM with the interpolated ones. Then, the optimization process in~\eqref{eq:erm} is reformulated as:
\begin{equation}
\label{eq:erm_mix}
\theta^{*} := \arg\min_{\theta \in \Theta} \mathbb{E}_{\{(x_i,y_i,d_i), (x_j, y_j,d_j)\}\sim \hat{P}} [\ell(f_{\theta}(x_{mix}), y_{mix})].
\end{equation}
Without additional selective augmentation strategies, vanilla mixup will regularize the model and reduce overfitting~\citep{zhang2020does}, allowing it to attain good in-distribution generalization. However, vanilla mixup may not be able to cancel out spurious correlations, causing the model to still fail at attaining good OOD generalization (see empirical comparisons in Section~\ref{sec:exp_compare_augmentation} and theoretical discussion in Section~\ref{sec:theory}). In LISA, we instead adopt a new strategy where mixup is only applied across specific domains or groups, which leans towards learning invariant 
predictors and thus better OOD performance. Specifically, the two kinds of selective augmentation strategies are presented as:

\noindent \textbf{Intra-label LISA (LISA-L): Interpolating samples with the same label.}
Intra-label LISA interpolates samples with the same label but different domains (i.e., $d_i \neq d_j$, $y_i=y_j$). As shown in Figure~\ref{fig:method_illustration}(a), this produces datapoints that have both domains partially present, effectively eliminating spurious correlations between domain and label in cases where the pair of domains correlate differently with the label. As a result, intra-label LISA should learn domain-invariant predictors for each class and thus achieve better OOD robustness. 

\noindent \textbf{Intra-domain LISA (LISA-D): Interpolating samples with the same domain.}
Supposing domain information is highly spuriously correlated with the label information, intra-domain LISA (Figure~\ref{fig:method_illustration}(b)) applies the interpolation strategy on samples with the same domain but different labels, i.e., $d_i=d_j$, $y_i\neq y_j$. Intuitively, even within the same domain, the model is supposed to generate different predicted labels since the interpolation ratio $\lambda$ is randomly sampled, corresponding to different labels $y_{mix}$. This causes the model to make predictions that are less dependent on the domain, again improving OOD robustness.

In this paper, we randomly perform intra-label or intra-domain LISA during the training process with probability $p_{sel}$ and $1-p_{sel}$, where $p_{sel}$ is treated as a hyperparameter and determined via cross-validation. \yao{Intuitively, the choice of $p_{sel}$ depends on the number of domains and the strength of the spurious correlations. Empirically, using intra-label LISA brings more benefits when there are more domains or when the the spurious correlations are not very strong. Intra-domain LISA benefits performance when domain information is highly spuriously correlated with the label. The pseudocode of LISA is in Algorithm~\ref{alg:ilsa}.}

\begin{algorithm}[H]
    \caption{Training Procedure of LISA}
    \label{alg:ilsa}
    \begin{algorithmic}[1]
    \REQUIRE Training data $\mathcal{D}$, step size $\eta$, learning rate $\gamma$, shape parameters $\alpha$, $\beta$ of Beta distribution
    \WHILE{not converge}
    \STATE Sample $\lambda \sim \mathrm{Beta}(\alpha, \beta)$
    \STATE Sample minibatch $B_1 \sim \mathcal{D}$
    \STATE Initialize $B_2 \leftarrow \{\}$
    \STATE Select strategy $s\sim Bernoulli(p_{sel})$
    \IF{$s$ is True} \comment{intra-label LISA}
    \FOR{$(x_i, y_i, d_i) \in B_1$}
        \STATE Randomly sample $(x_j, y_j, d_j) \sim \{(x,y,d) \in \mathcal{D}\}$ which satisfies $(y_i=y_j)$ and $(d_i\neq d_j)$. 
        \STATE Put $(x_j, y_j, d_j)$ into $B_2$. 
    \ENDFOR
    \ELSE \comment{intra-domain LISA}
    \FOR{$(x_i, y_i, d_i) \in B_1$}
        \STATE Randomly sample $(x_j, y_j, d_j) \sim \{(x,y,d) \in \mathcal{D}\}$ which satisfies $(y_i\neq y_j)$ and $(d_i= d_j)$. 
        \STATE Put $(x_j, y_j, d_j)$ into $B_2$. 
    \ENDFOR
    \ENDIF
    \STATE Update $\theta$ with data $\lambda B_1 + (1-\lambda) B_2$ with learning rate $\gamma$. 
    \ENDWHILE
    \end{algorithmic}
\end{algorithm}
\section{Experiments}
\label{sec:experiments}
\begin{table*}[h]
\small
\caption{Dataset Statistics for Subpopulation Shifts. All datasets are binary classification tasks and we use the worst group accuracy as the evaluation metric.}
\label{tab:subpopulation_data}
\begin{center}
\begin{tabular}{l|ccc}
\toprule
Datasets  & Domains  & Model Architecture & Class Information \\\midrule
CMNIST & 2 digit colors  & ResNet-50 & digit (0,1,2,3,4) v.s. (5,6,7,8,9)\\
Waterbirds & 2 backgrounds  & ResNet-50 & waterbirds v.s. landbirds\\
CelebA & 2 hair colors & ResNet-50 & man v.s. women\\
CivilComments & 8 demographic
identities & DistilBERT-uncased & toxic v.s. non-toxic \\\bottomrule
\end{tabular}
\end{center}
\vspace{-1.5em}
\end{table*}
In this section, we conduct comprehensive experiments to evaluate the effectiveness of LISA. Specifically, we aim to answer the following questions: \textbf{Q1}: Compared to prior methods, can LISA improve robustness to subpopulation shifts and domain shifts (Section~\ref{sec:exp_main_sub} and Section~\ref{sec:exp_main_domain})? \textbf{Q2}: Which aspects of LISA are the most important for improving robustness (Section~\ref{sec:exp_compare_augmentation})? \textbf{Q3}: Does LISA successfully produce more invariant predictors (Section~\ref{sec:invariance})? \textbf{Q4}: How does LISA perform with varying degrees of distribution shifts (Section~\ref{sec:exp_degree})? 

To answer Q1, we compare to ERM, IRM~\citep{arjovsky2019invariant}, IB-IRM~\citep{ahuja2021invariance}, V-REx~\citep{krueger2021out}, CORAL~\citep{li2018domain}, DRNN~\citep{ganin2015unsupervised}, GroupDRO~\citep{sagawa2019distributionally}, DomainMix~\citep{xu2020adversarial}, and Fish~\citep{shi2021gradient}. Upweighting (UW) is particularly suitable for subpopulation shifts, so we also use it for comparison. We adopt the same model architectures for all approaches. The strategy selection probability $p_{sel}$ is selected via cross-validation.

\subsection{Evaluating Robustness to Subpopulation Shifts}
\label{sec:exp_main_sub}
\textbf{Evaluation Protocol.} In subpopulation shifts, we evaluate the performance on four binary classification datasets, including Colored MNIST (CMNIST), Waterbirds~\citep{sagawa2019distributionally}, CelebA~\citep{liu2015faceattributes}, and Civilcomments~\citep{borkan2019nuanced}. We detail the data descriptions of subpopulation shifts in Appendix~\ref{sec:app_sub_data} and report the detailed data statistics in Table~\ref{tab:subpopulation_data}, covering domain information, model architecture, and class information. Following~\citet{sagawa2019distributionally}, in subpopulation shifts, we use the worst-group accuracy to evaluate the performance of all approaches. In these datasets, the domain information is highly spurious correlated with the label information. For example, as suggested in Figure~\ref{fig:method_illustration}, 80\% images in the CMNIST dataset have the same color in each specific class, i.e., green color for label [1, 0] and red color for label [0, 1]. 

In CMNIST, Waterbirds, and CelebA, we find that $p_{sel}=0.5$ works well for choosing selective augmentation strategies, while in CivilComments, we set $p_{sel}$ as $1.0$ . This is not surprising because it might be more beneficial to use intra-label LISA more often to eliminate domain effects when there are more domains, i.e., eight domains in CivilComments v.s. two domains in others. The rest hyperparameter settings and training details are listed in Appendix~\ref{sec:app_sub_training}.

\begin{table*}[h]
\small
\caption{Results of subpopulation shifts. Here, we show the average and worst group accuracy. We repeat the experiments three times and put full results with standard deviation in Table~\ref{tab:subpopulation_main_full}.}
\label{tab:subpopulation_main}
\begin{center}
\begin{tabular}{l|cc|cc|cc|cc}
\toprule
\multirow{2}{*}{} & \multicolumn{2}{c|}{CMNIST} & \multicolumn{2}{c|}{Waterbirds} & \multicolumn{2}{c|}{CelebA} & \multicolumn{2}{c}{CivilComments} \\
& Avg. & Worst  & Avg. & Worst  & Avg. & Worst & Avg. & Worst \\\midrule
ERM & 27.8\% & 0.0\% & 97.0\% & 63.7\% & 94.9\% & 47.8\% & 92.2\% & 56.0\% \\
UW  & 72.2\% & 66.0\% &  95.1\% & 88.0\% & 92.9\% & 83.3\% & 89.8\% & 69.2\% \\
IRM & 72.1\% & 70.3\% & 87.5\% & 75.6\% & 94.0\% & 77.8\% & 88.8\% & 66.3\% \\
IB-IRM & 72.2\% & 70.7\% & 88.5\% & 76.5\% & 93.6\% & 85.0\% & 89.1\% & 65.3\%\\
V-REx & 71.7\% & 70.2\% & 88.0\% & 73.6\% & 92.2\% & 86.7\% &  90.2\% & 64.9\% \\
CORAL & 71.8\% & 69.5\% & 90.3\% & 79.8\% & 93.8\% & 76.9\% & 88.7\% & 65.6\% \\
GroupDRO & 72.3\% & 68.6\% & 91.8\% & \textbf{90.6\%} & 92.1\% & 87.2\% & 89.9\% & 70.0\%  \\
DomainMix & 51.4\% & 48.0\% & 76.4\% & 53.0\% & 93.4\% & 65.6\% & 90.9\% & 63.6\%\\
Fish & 46.9\% & 35.6\% & 85.6\% & 64.0\% & 93.1\% & 61.2\% &  89.8\% & 71.1\% \\
\midrule
\textbf{LISA (ours)} & 74.0\% & \textbf{73.3\%} & 91.8\% & 89.2\% & 92.4\% & \textbf{89.3\%} & 89.2\% &  \textbf{72.6\%} \\
\bottomrule
\end{tabular}
\end{center}
\vspace{-2em}
\end{table*}

\begin{table*}[h]
\small
\caption{Main domain shifts results. LISA outperforms prior methods on all five datasets. Following the instructions of~\citet{koh2021wilds}, we report the performance of Camelyon17 over 10 different seeds and the results of other datasets are obtained over 3 different seeds.}
\label{tab:res_domain}
\begin{center}
\begin{tabular}{l|cccccc}
\toprule
\multirow{2}{*}{}  & Camelyon17 & FMoW & RxRx1 & Amazon &  MetaShift\\\cmidrule{2-6}
& Avg. Acc. & Worst Acc. & Avg. Acc. & 10-th Per. Acc. & Worst Acc.\\\midrule
ERM & 70.3 $\pm$ 6.4\% & 32.3 $\pm$ 1.25\%  & 29.9 $\pm$ 0.4\% & 53.8 $\pm$ 0.8\% & 52.1 $\pm$ 0.4\% \\
IRM  & 64.2 $\pm$ 8.1\% & 30.0 $\pm$ 1.37\% & 8.2 $\pm$ 1.1\% &  52.4 $\pm$ 0.8\% & 51.8 $\pm$ 0.8\% \\
IB-IRM & 68.9 $\pm$ 6.1\% & 28.4 $\pm$ 0.90\% & 6.4 $\pm$ 0.6\% & 53.8 $\pm$ 0.7\% & 52.3 $\pm$ 1.0\% \\
V-REx & 71.5 $\pm$ 8.3\% & 27.2 $\pm$ 0.78\% & 7.5 $\pm$ 0.8\% & 53.3 $\pm$ 0.0\% & 51.6 $\pm$ 1.8\% \\
CORAL & 59.5 $\pm$ 7.7\% & 31.7 $\pm$ 1.24\% & 28.4 $\pm$ 0.3\% & 52.9 $\pm$ 0.8\% & 47.6 $\pm$ 1.9\%\\
GroupDRO & 68.4 $\pm$ 7.3\% & 30.8 $\pm$ 0.81\% & 23.0 $\pm$ 0.3\% & 53.3 $\pm$ 0.0\% & 51.9 $\pm$ 0.7\% \\
DomainMix & 69.7 $\pm$ 5.5\% & 34.2 $\pm$ 0.76\% & 30.8 $\pm$ 0.4\% & 53.3 $\pm$ 0.0\% & 51.3 $\pm$ 0.5\% \\
Fish & 74.7 $\pm$ 7.1\% & 34.6 $\pm$ 0.18\% & 10.1 $\pm$ 1.5\%  & 53.3 $\pm$ 0.0\% & 49.2 $\pm$ 2.1\% \\\midrule
\textbf{LISA (ours)} & \textbf{77.1 $\pm$ 6.5\%} & \textbf{35.5 $\pm$ 0.65\%} & \textbf{31.9 $\pm$ 0.8\%} & \textbf{54.7 $\pm$ 0.0\%} & \textbf{54.2 $\pm$ 0.7\%} \\
\bottomrule
\end{tabular}
\end{center}
\vspace{-2em}
\end{table*}

\textbf{Results.} 
In Table~\ref{tab:subpopulation_main}, we report the overall performance of LISA and other methods. According to Table~\ref{tab:subpopulation_main}, we \yao{observe that the performance of approaches that learn invariant predictors with explicit regularizers (e.g., IRM, IB-IRM, V-REx) is not consistent across datasets. For example, IRM and V-REx outperform UW on CMNIST, but they fail to achieve better performance than UW on Waterbirds. The results corroborate our hypothesis that designing widely effective regularizers is challenging, and that inappropriate regularizers may even hurt the performance. LISA instead consistently outperforms other invariant learning methods (e.g., IRM, IB-IRM, V-REx, CORAL, DomainMix, Fish) in all datasets. LISA further shows the best performance on CMNIST, CelebA, and CivilComments. In Waterbirds, it is slightly worse than GroupDRO, but the performance is comparable. These results demonstrate the effectiveness of LISA in improving robustness to subpopulation shifts.}

\begin{figure}[h]
    \centering
    \small
    \includegraphics[width=0.32\textwidth]{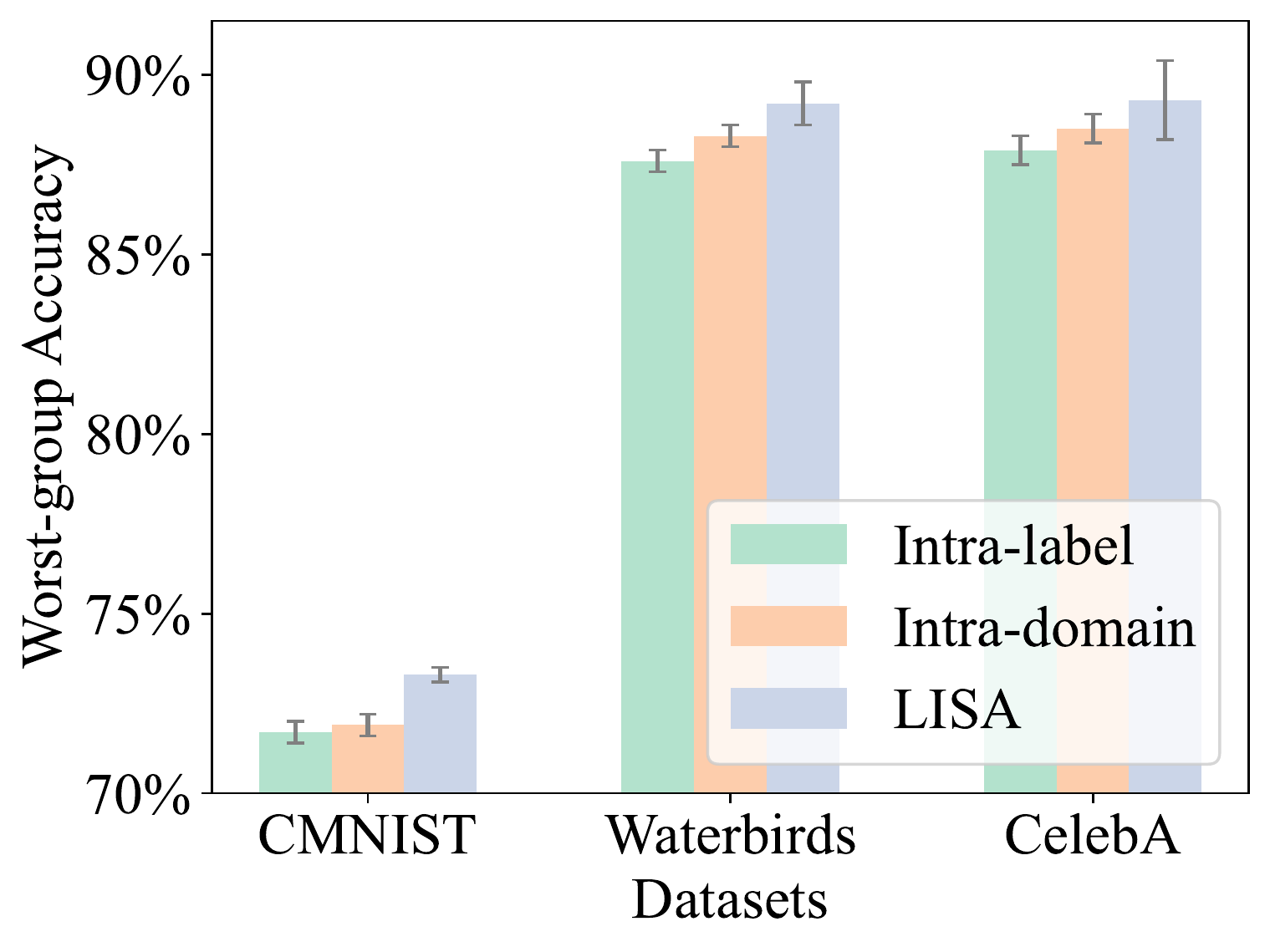}
    \vspace{-0.8em}
    \caption{Effects of intra-label and intra-domain LISA in CMNIST, Waterbirds and CelebA. The experiments are repeated three times with different seeds.}
    \label{fig:intraonly}
\end{figure}

\textbf{Effects of Intra-label and Intra-domain LISA.} For CMNIST, Waterbirds and CelebA, both intra-label and intra-domain LISA are used (i.e., $p_{sel}=0.5$), we illustrate the separate results in Figure~\ref{fig:intraonly} and observe that both variants contribute to the final performance. In addition, intra-domain LISA performs slightly better than intra-label LISA, corroborating our assumption that intra-domain LISA benefits more when domain information is highly spuriously correlated with the label (see the discussion of the strength of spurious correlation in Appendix~\ref{sec:app_spurious_strength}). 

\subsection{Evaluating Robustness to Domain Shifts}
\label{sec:exp_main_domain}
\begin{table*}[h]
\caption{Dataset Statistics for Domain Shifts.}
\vspace{0.5em}
\small
\label{tab:domian_data}
\begin{center}
\begin{tabular}{l|cccc }
\toprule
Datasets  & Domains & Metric & Base Model & Num. of classes \\\midrule
Camelyon17 & 5 hospitals & Avg. Acc. & DenseNet-121 & 2\\
FMoW & 16 years x 5 regions & Worst-group Acc. & DenseNet-121 & 62\\
RxRx1 & 51 experimental batches & Avg. Acc. & ResNet-50 & 1,139\\
Amazon & 7,676 reviewers & 10th Percentile Acc. & DistilBERT-uncased & 5\\
MetaShift & 4 backgrounds & Worst-group Acc. & ResNet-50 & 2 \\
\bottomrule
\end{tabular}
\end{center}
\vspace{-2em}
\end{table*}
\textbf{Experimental Setup.} In domain shifts, we study five datasets. Four of them (Camelyon17, FMoW, RxRx1, and Amazon) are selected from WILDS~\citep{koh2021wilds}, covering natural distribution shifts across diverse domains (e.g., health, language, and vision). Besides the WILDS data, we also apply LISA on the MetaShift datasets~\citep{metadataset}, constructed using the real-world images and natural heterogeneity of Visual Genome~\citep{krishnavisualgenome}. We summarize these datasets in Table~\ref{tab:domian_data}, including domain information, evaluation metric, model architecture, and the number of classes. Detailed dataset descriptions and other training details are discussed in Appendix~\ref{sec:app_domain_data} and~\ref{sec:app_domain_training}, respectively.

\yao{The strategy selection probability $p_{sel}$ is set as $1.0$ for these domain shifts datasets, i.e., only intra-label LISA is used. Additionally, we only interpolate samples with the same labels without considering the domain information in Camelyon17, FMoW, and RxRx1, which empirically leads to the best performance. One potential reason is that the spurious correlations between labels and domains are not very strong in datasets with natural domain shifts under the existing domain partitions. Here, to evaluate the strength of spurious correlation, we adopt Cramér's V~\citep{cramer2016mathematical} (see the detailed definition in Appendix~\ref{sec:app_spurious_strength}) to measure the association between the domain set $\mathcal{D}$ and the label set $\mathcal{Y}$, where the results are reported in Table~\ref{tab:spurious_strength} of Appendix~\ref{sec:app_spurious_strength}. The Cramér's V values in Camelyon17, FMoW, and RxRx1 are significantly smaller than other datasets, indicating relatively weak spurious correlations. Under this setting, enlarging the interpolation scope by directly interpolating samples within the same class regardless of existing domain information may bring more benefits.}

\yao{\textbf{Results.}
We report the results of domain shifts in Table~\ref{tab:res_domain}, where full results that include validation performance and other metrics are listed in Appendix~\ref{sec:app_domain_full_results}. Aligning with the observation in subpopulation shifts, the performance of prior regularization-based invariant predictor learning methods (e.g., IRM, IB-IRM, V-REx) is still unstable across different datasets. For example, V-REx outperforms ERM on Camelyon17, while it fails in RxRx1. However, LISA consistently outperforms all these methods on five datasets regardless of the model architecture and data types (i.e., image or text), indicating its effectiveness in improving robustness to domain shifts with selective augmentation.}

\subsection{Are the Performance Gains of LISA from Data Augmentation?}

\begin{table*}[h]
\small
\caption{Compared LISA with substitute mixup strategies in domain shifts.}
\vspace{0.5em}
\label{tab:domain_ablation}
\begin{center}
\begin{tabular}{l|cccccc}
\toprule
\multirow{2}{*}{}  & Camelyon17 & FMoW & RxRx1 & Amazon &  MetaShift\\\cmidrule{2-6}
& Avg. Acc. & Worst Acc. & Avg. Acc. & 10-th Per. Acc. & Worst Acc.\\\midrule
ERM & 70.3 $\pm$ 6.4\% & 32.8 $\pm$ 0.45\%  & 29.9 $\pm$ 0.4\% & 53.8 $\pm$ 0.8\% & 52.1 $\pm$ 0.4\% \\
Vanilla mixup & 71.2 $\pm$ 5.3\% & 34.2 $\pm$ 0.45\% & 26.5 $\pm$ 0.5\% & 53.3 $\pm$ 0.0\% & 51.3 $\pm$ 0.7\% \\
In-group mixup & 75.5 $\pm$ 6.7\% & 32.2 $\pm$ 1.18\% & 24.4 $\pm$ 0.2\% & 53.8 $\pm$ 0.6\% & 52.7 $\pm$ 0.5\%\\\midrule
\textbf{LISA (ours)} & \textbf{77.1 $\pm$ 6.5\%} & \textbf{35.5 $\pm$ 0.65\%} & \textbf{31.9 $\pm$ 0.8\%} & \textbf{54.7 $\pm$ 0.0\%} & \textbf{54.2 $\pm$ 0.7\%} \\
\bottomrule
\end{tabular}
\end{center}
\vspace{-1em}
\end{table*}

\begin{table*}[h]
\small
\caption{Compared LISA with substitute mixup strategies in subpopulation shifts. UW represents upweighting. Full results with standard deviation is listed in Table~\ref{tab:subpopulation_ablation_full}.}
\label{tab:subpopulation_ablation}
\begin{center}
\begin{tabular}{l|cc|cc|cc|cc}
\toprule
\multirow{2}{*}{} & \multicolumn{2}{c|}{CMNIST} & \multicolumn{2}{c|}{Waterbirds} & \multicolumn{2}{c|}{CelebA} & \multicolumn{2}{c}{CivilComments}\\
& Avg. & Worst  & Avg. & Worst  & Avg. & Worst & Avg. & Worst \\\midrule
ERM & 27.8\% & 0.0\% & 97.0\% & 63.7\% & 94.9\% & 47.8\% & 92.2\% & 56.0\% \\
Vanilla mixup & 32.6\% & 3.1\% &  81.0\% & 56.2\% & 95.8\% & 46.4\% & 90.8\% & 67.2\% \\
Vanilla mixup + UW & 72.2\% & 71.8\% & 92.1\% & 85.6\% & 91.5\% & 88.0\% & 87.8\% & 66.1\% \\
In-group mixup & 33.6\% & 24.0\% & 88.7\% & 68.0\% & 95.2\% & 58.3\% & 90.8\% & 69.2\% \\
In-group mixup + UW & 72.6\% & 71.6\% & 91.4\% & 87.1\%  & 92.4\%  & 87.8\% & 84.8\% & 69.3\%  \\
\midrule
\textbf{LISA (ours)} & 74.0\% & \textbf{73.3\%} & 91.8\% & \textbf{89.2\%} & 92.4\% & \textbf{89.3\%} & 89.2\% & \textbf{72.6\%} \\
\bottomrule
\end{tabular}
\end{center}
\vspace{-1em}
\end{table*}
\begin{table*}[h]
\small
\caption{Results of the analysis of learned invariant predictors. Accuracy of domain prediction ($\mathrm{IP}_{adp}$) and pairwise divergence of prediction among all domains ($\mathrm{IP}_{kl}$) are used to measure the invariance. Smaller values denote stronger invariance.}
\label{tab:prediction_invariance}
\begin{center}
\setlength{\tabcolsep}{1.3mm}{
\begin{tabular}{l|c|c|c|c|c|c|c|c}
\toprule
& \multicolumn{4}{c|}{$\mathrm{IP}_{adp} \downarrow$} & \multicolumn{4}{c}{$\mathrm{IP}_{kl} \downarrow$} \\\cmidrule{2-9}
  & CMNIST & Waterbirds & Camelyon17  & MetaShift & CMNIST & Waterbirds  & Camelyon17 & MetaShift\\\midrule
ERM & 82.85\% & 94.99\% &  49.43\% & 67.98\%  & 6.286 & 1.888 & 1.536 & 1.205 \\
Vanilla mixup & 92.34\% & 94.49\% &  52.79\% & 69.36\% & 4.737 & 2.912 & 0.790 & 1.171\\
IRM & 69.42\% & 95.12\% & 47.96\% & 67.59\% & 7.755 & 1.122 & 0.875 & 1.148 \\
IB-IRM & 74.72\% & 94.78\% &  48.37\% & 67.39\% & 1.004 & 3.563 & 0.756 & 1.115 \\
V-REx & 63.58\% & 93.32\%  & 61.38\% & 68.38\% & 3.190 & 3.791 & 1.281 & 1.094 \\
\midrule
\textbf{LISA (ours)} & \textbf{58.42\%} & \textbf{90.28\%} & \textbf{45.15\%} & \textbf{66.01\%}  &  \textbf{0.567} & \textbf{0.134} & \textbf{0.723}  & \textbf{1.001} \\ 
\bottomrule
\end{tabular}
}
\vspace{-2em}
\end{center}
\end{table*}

\label{sec:exp_compare_augmentation}
In LISA, we apply selective augmentation strategies on samples either with the same label but different domains or with the same domain but different labels. Here, we explore two substitute interpolation strategies: 
\begin{itemize}[leftmargin=*]
    \item \emph{Vanilla mixup}: in Vanilla mixup, we do not add any constraints on the sample selection, i.e., the mixup is performed on any pairs of samples.
    \item \emph{In-group mixup}: this strategy applies data interpolation on samples with the same labels and from the same domains. 
\end{itemize}
Notice that all substitute interpolation strategies use the same variant of mixup as LISA (e.g., mixup/CutMix). Finally, as upweighting (UW) small groups significantly improves performance in subpopulation shifts, we evaluate UW combined with Vanilla/In-group mixup.

The results of substitute interpolation strategies on domain shifts and subpopulation shifts are in Table~\ref{tab:domain_ablation} and Table~\ref{tab:subpopulation_ablation}, respectively. Furthermore, we also conduct experiments on datasets without spurious correlation in Table~\ref{tab:app_no_spurious} of Appendix~\ref{sec:app_no_spurious}. From the results, we make the following three key observations. \emph{First}, compared with Vanilla mixup, the performance of LISA verifies that selective data interpolation indeed improve the out-of-distribution robustness by canceling out the spurious correlations and encouraging learning invariant predictors rather than simple data augmentation. These findings are further strengthened by the results in Table~\ref{tab:app_no_spurious} of Appendix~\ref{sec:app_no_spurious}, where Vanilla mixup outperforms LISA and ERM without spurious correlations but LISA achieves the best performance with spurious correlations. \emph{Second}, the superiority of LISA over In-group mixup verifies that only interpolating samples within each group is incapable of eliminating out the spurious information, where In-group mixup still performs the role of data augmentation. \emph{Third}, though incorporating UW significantly improves the performance of Vanilla mixup and In-group mixup in subpopulation shifts, LISA still achieves larger benefits than these enhanced substitute strategies, demonstrating its stronger power in improving OOD robustness.

\subsection{Does LISA Lead to More Invariant Predictors?}
\label{sec:invariance}

\yao{We further analyze the model invariance learned by LISA. Specifically, for each sample {($x_i$, $y_i$, $d$)} in domain $d$, we denote the unscaled output (i.e., logits) of the model as $g_{i,d}$. We use two metrics to measure the invariance (see Appendix~\ref{sec:app_additional_predictor_invariance} for additional metrics and the corresponding results): }
\vspace{-1em}
\begin{itemize}[leftmargin=*]
    \item \yao{\textbf{Accuracy of domain prediction} ($\mathrm{IP}_{adp}$). In the first metric, we use the unscaled output to predict the domain. Concretely, the entire dataset is re-split into training, validation, and test sets, where logits are used as features and labels represent the corresponding domain ID. A logistic regression model is trained to predict the domain.}
    \vspace{-0.5em}
    \item \yao{\textbf{Pairwise divergence of prediction} ($\mathrm{IP}_{kl}$). We calculate the KL divergence of the distribution of logits among all domains, where kernel density estimation is used to estimate the probability density function $P(g_{d}^y)$ of logits from domain $d$ with label $y$. The pairwise divergence of the predictions is defined as \begin{small}$\frac{1}{|\mathcal{Y}||\mathcal{D}|^2}\sum_{y\in \mathcal{Y}}\sum_{d',d\in \mathcal{D}}\mathrm{KL}(P(g_D^y\mid D=d)|P(g_D^y\mid D=d'))$\end{small}.}
\end{itemize}
\vspace{-1em}

\yao{Small values of $\mathrm{IP}_{adp}$ and $\mathrm{IP}_{kl}$ represent strong function-level invariance. In Table~\ref{tab:prediction_invariance}, we report the results of LISA and other approaches on CMNIST, Waterbirds, Camelyon17 and MetaShift. The results verify that LISA learns predictors with greater domain invariance. Besides having more invariant predictors, we observe that LISA also leads to more invariant representations, as detailed in Appendix~\ref{app:sec_representation_invariance}.}

\subsection{Effect of the Degree of Distribution Shifts}
\label{sec:exp_degree}
We investigate the performance of LISA with respect to the degree of distribution shifts. Here, we use MetaShift to evaluate performance, where the distance between training and test domains is measured as the node similarity on a meta-graph~\citep{metadataset}. To vary the distance between training and test domains, we change the backgrounds of training objects (see full experimental details in Appendix~\ref{sec:app_domain_data}). The performance with varied distances is illustrated in Table~\ref{tab:vary_distance}, where the top four best methods (i.e., ERM, IRM, IB-IRM, GroupDRO) are reported for comparison. We observe that LISA consistently outperforms other methods under all scenarios. Another interesting finding is that LISA achieves more substantial improvements with the increases of distance. A potential reason is that the effects of eliminating domain information is more obvious when there is a larger distance between training and test domains. 

\begin{table}[h]
\small
\vspace{-0.5em}
\caption{Effects of the degree of distribution shifts w.r.t. the performance on the MetaShift benchmark. Distance represents the distribution distance between training and test domains. Best B/L represents best baseline.}
\label{tab:vary_distance}
\begin{center}
\setlength{\tabcolsep}{1.mm}{
\begin{tabular}{l|cccc}
\toprule
Distance & 0.44 & 0.71 & 1.12 & 1.43 \\\midrule
ERM & 80.1\% & 68.4\% & 52.1\% & 33.2\% \\
IRM & 79.5\% & 67.4\% & 51.8\% & 32.0\% \\
IB-IRM & 79.7\% & 66.9\% & 52.3\% & 33.6\% \\
GroupDRO & 77.0\% & 68.9\% & 51.9\% & 34.2\% \\
LISA (ours) & \textbf{81.3\%} & \textbf{69.7\%} & \textbf{54.2\%} & \textbf{37.5\%} \\\midrule
LISA v.s. Best B/L & \textcolor{Green}{+1.5\%} & \textcolor{Green}{+1.2\%} & \textcolor{Green}{+3.6\%} & \textcolor{Green}{+9.6\%}\\
\bottomrule
\end{tabular}
}
\end{center}
\end{table}

\section{Theoretical Analysis}
\label{sec:theory}
In this section, we provide some theoretical understandings that explain several of the empirical phenomena from the previous experiments and theoretically compare the worst-group errors of three methods: the proposed LISA, ERM, and vanilla mixup. Specifically, we consider a Gaussian mixture model with subpopulation and domain shifts, which has been widely adopted in theory to shed light upon complex machine learning phenomenon such as in \cite{montanari2019generalization,zhang2021and, liu2021self}. We   note here that despite the popularity of mixup in practice, the theoretical analysis of how mixup (w/ or w/o the selective augmentation strategies) affects the misclassification error is still largely unexplored in the literature even in the simple models. As discussed in Section~\ref{sec:prelim}, here, we define \begin{small}$y\in\{0,1\}$\end{small} as the label, and \begin{small}$d\in\{R,G\}$\end{small} as the domain information. For \begin{small}$y\in\{0,1\}$\end{small} and \begin{small}$d\in\{R,G\}$\end{small}, we consider the following model:
\begin{equation}
\small
\label{m1}
    x_i|y_i=y,d_i=d\sim N(\mu^{(y,d)},\Sigma^{(d)}), i=1,\dots,n^{(y,d)},
\end{equation}
where \begin{small}$\mu^{(y,d)}\in\R^p$\end{small} is the conditional mean vector and \begin{small}$\Sigma^{(d)}\in\R^{p\times p}$\end{small} is the covariance matrix. Let \begin{small}$n=\sum_{y\in\{0,1\},d\in\{R,G\}}n^{(y,d)}$\end{small}. 
Let \begin{small}$\pi^{(y,d)}=\mathbbm{P}(y_i=y,d_i=d)$\end{small}, \begin{small}$\pi^{(y)}=\mathbbm{P}(y_i=y)$\end{small}, and \begin{small}$\pi^{(d)}=\mathbbm{P}(d_i=d)$\end{small}.

To account for the spurious correlation brought by domains, we consider \begin{small}$\mu^{(y,R)}\neq \mu^{(y,G)}$\end{small} in general for \begin{small}$y\in\{0,1\}$\end{small} and the imbalanced case where \begin{small}$\pi^{(0,R)}, \pi^{(1,G)}<1/4$\end{small}. Moreover, we assume there exists some invariance across different domains. Specifically, we assume 
\begin{equation*}
   \resizebox{\hsize}{!}{$\mu^{(1,R)}-\mu^{(0,R)}= \mu^{(1,G)}-\mu^{(0,G)}:=\Delta~~\text{and}~~\Sigma^{(G)}=\Sigma^{(R)}:=\Sigma$}.
\end{equation*}
According to Fisher's linear discriminant analysis \citep{anderson1962introduction,tony2019high,cai2021convex}, the optimal classification rule is linear with slope \begin{small}$\Sig^{-1}\Delta$\end{small}. The assumption above implies that \begin{small}$(\Sig^{-1}\Delta)^\top x$\end{small} is the (unknown) invariant prediction rule for model \eqref{m1}. 

Suppose we use some method \begin{small}$A$\end{small} and obtain a linear classifier \begin{small}$x^Tb+b_0>0$\end{small} from a training data, we will apply it to a test data and compute the worst-group misclassification error, where the mis-classification error for domain $d$ and class $y$ is \begin{small}$E^{(y,d)}(b,b_0):=  \mathbbm{P}(\mathbbm{1}(x_i^Tb+b_0>\frac{1}{2})\neq y|d_i=d,y_i=y)$\end{small}, 
and we denote the worst-group error with the method $A$ as
\begin{equation}
\nonumber
\small
E^{(wst)}_A= \max_{d\in\{R,G\},y\in\{0,1\}}E^{(y,d)}(b_A,b_{0,A}),
\end{equation}
where \begin{small}$b_A$\end{small} and \begin{small}$b_{0,A}$\end{small} are the slope and intercept based on the method \begin{small}$A$\end{small}. Specifically, \begin{small}$A=\textup{ERM}$\end{small} denotes the ERM method (by minimizing the sum of squares loss on the training data altogether), \begin{small}$A=\textup{mix}$\end{small} denotes the vanilla mixup method (without any selective augmentation strategy), and \begin{small}$A=\textup{LISA}$\end{small} denotes the mixup strategy for LISA. We also denote its finite sample version by \begin{small}$\hat E_A^{(wst)}$\end{small}.

Let \begin{small}$\widetilde{\Delta}=\E[x_i|y_i=1]-\E[x_i|y_i=0]$\end{small} denote the marginal difference and \begin{small}$\xi=\frac{\Delta^T\Sig^{-1}\widetilde{\Delta}}{\|\Delta\|_{\Sig}\|\widetilde{\Delta}\|_{\Sig}}$\end{small} denote the correlation operator between the domain-specific difference \begin{small}$\Delta$\end{small} and the marginal difference \begin{small}$\widetilde{\Delta}$\end{small} with respect to \begin{small}$\Sig$\end{small}. We see that smaller \begin{small}$\xi$\end{small} indicates larger discrepancy between the marginal difference and the domain-specific difference and therefore implies stronger spurious correlation between the domains and labels. We present the following theorem showing that our proposed LISA algorithm outperforms the ERM and vanilla mixup in the subpopulation shifts setting. 

\begin{theorem}[Error comparison with subpopulation shifts]
\label{thm1}

Consider $n$ independent samples generated from model (\ref{m1}),  \begin{small}$\pi^{(R)}=\pi^{(1)}=1/2$\end{small}, \begin{small}$\pi^{(0,R)}=\pi^{(1,G)}=\alpha<1/4$\end{small}, \begin{small}$\max_{y,d}\|\mu^{(y,d)}\|_2\leq C$\end{small}, and \begin{small}$\Sig$\end{small} is positive definite. Suppose \begin{small}$(\xi,\alpha)$\end{small} satisfies that \begin{small}$\xi < \min\{\frac{\|\widetilde{\Delta}\|_{\Sig}}{\|\Delta\|_{\Sig}},\frac{\|\Delta\|_{\Sig}}{\|\widetilde{\Delta}\|_{\Sig}}\}-C\alpha$\end{small}
 for some large enough constant \begin{small}$C$\end{small} and \begin{small}$\E[\lam_i^2]/\max\{var(\lam_i),1/4\}\geq \|\widetilde{\Delta}\|_{\Sig}^2+\|\widetilde{\Delta}\|_{\Sig}\|\Delta\|_{\Sig}$\end{small}.
Then for any \begin{small}$p_{sel}\in[0,1]$\end{small},
\begin{small}
$$\widehat{E}_{\textup{LISA}}^{(wst)}< \min\{\widehat{E}_{\textup{ERM}}^{(wst)},\widehat{E}_{\textup{mix}}^{(wst)}\}+O_P\left(\frac{p\log n}{n}+\frac{p}{\alpha n}\right).$$
\end{small}
\end{theorem}
In Theorem \ref{thm1}, $\lambda_i$ is the random mixup coefficient for the $i$-th sample. If $\lambda_i=\lambda$ are the same for all the samples in a mini-batch, the results still hold. 
Theorem \ref{thm1} implies that when \begin{small}$\xi$\end{small} is small (indicating that the domain has strong spurious correlation with the label) and \begin{small}$p=o(\alpha n)$\end{small}, the worst-group classification errors of LISA are asymptotically smaller than that of ERM and vanilla mixup. {\color{black} In fact, our analysis shows that LISA yields a classification rule closer to the invariant classification rules by leveraging the domain information.}

In the next theorem, we present the mis-classification error comparisons with domain shifts. That is, consider samples from a new unseen domain: 
\begin{equation}
\small
    x_i^{(0,*)}\sim N(\mu^{(0,*)},\Sigma),~~x_i^{(1,*)}\sim N(\mu^{(1,*)},\Sigma).
\end{equation}
  
Let \begin{small}$\widetilde{\Delta}^*=2(\mu^{(0,*)}-\E[x_i])$\end{small}, where \begin{small}$\E[x_i]$\end{small} is the mean of the training distribution, and assume \begin{small}$\mu^{(1,*)}-\mu^{(0,*)}=\Delta$\end{small}. Let \begin{small}$\xi^*=\frac{\widetilde{\Delta}^T\Sig^{-1}\widetilde{\Delta}^*}{\|\widetilde{\Delta}\|_{\Sig}\|\Delta\|_{\Sig}}$\end{small} and \begin{small}$\gamma=\frac{\Delta^T\Sig^{-1}\widetilde{\Delta}^*}{\|\widetilde{\Delta}\|_{\Sig}\|\Delta\|_{\Sig}}$\end{small} denote the correlation for \begin{small}$(\widetilde{\Delta}^*,\widetilde{\Delta})$\end{small} and for \begin{small}$(\widetilde{\Delta}^*,\Delta)$\end{small}, respectively, with respect to \begin{small}$\Sig^{-1}$\end{small}. 
Let \begin{small}$E^{(wst*)}_A= \max_{y\in\{0,1\}}E^{(y,*)}(b_A,b_{0,A})$\end{small} and its sample version be \begin{small}$\hat E^{(wst*)}_A$\end{small}.

\begin{theorem}[Error comparison with domain shifts]
\label{thm3}
Suppose $n$ samples are independently generated from model (\ref{m1}), \begin{small}$\pi^{(R)}=\pi^{(1)}=1/2,\pi^{(0,R)}=\pi^{(1,G)}=\alpha<1/4$\end{small},  \begin{small}$\max_{y,d}\|\mu^{(y,d)}\|_2\leq C$\end{small} and \begin{small}$\Sig$\end{small} is positive definite.
Suppose that \begin{small}$(\xi,\xi^*,\gamma)$\end{small} satisfy that \begin{small}$0\leq \xi^*\leq \gamma\xi$\end{small} and \begin{small}$\xi < \min\{\frac{\gamma}{2}\frac{\|\widetilde{\Delta}\|_{\Sig}}{\|\Delta\|_{\Sig}},\frac{\|\Delta\|_{\Sig}}{\|\tilde{\Delta}\|_{\Sig}}\}-C\alpha$\end{small}
 for some large enough constant $C$ and \begin{small}$\E[\lam_i^2]/\max\{var(\lam_i),1/4\}\geq \|\widetilde{\Delta}\|_{\Sig}^2+\|\widetilde{\Delta}\|_{\Sig}\|\Delta\|_{\Sig}$\end{small}.
Then for any \begin{small}$p_{sel}\in[0,1]$\end{small},
\begin{small}$$\widehat{E}_{\textup{LISA}}^{(wst*)}< \min\{\widehat{E}_{\textup{ERM}}^{(wst*)},\widehat{E}_{\textup{mix}}^{(wst*)}\}+O_P\left(\frac{p\log n}{n}+\frac{p}{\alpha n}\right).$$
\end{small}
\end{theorem}

Similar to Theorem~\ref{thm1}, this result shows that when domain has strong spurious correlation with the label (corresponding to small $\xi$), such a spurious correlation leads to the downgraded performance of ERM and vanilla mixup, while our proposed LISA method is able to mitigate such an issue by selective data interpolation. Proofs of Theorem~\ref{thm1} and Theorem~\ref{thm3} are provided in Appendix~\ref{sec:app_proof}.
\section{Related Work and Discussion}
In this paper, we focus on improving the robustness of machine learning models to subpopulation shifts and domain shifts. Here, we discuss related approaches from the following three categories:

\textbf{Learning Invariant Representations.} Motivated by unsupervised domain adaptation~\citep{ben2010theory,ganin2016domain}, the first category of works learns invariant representations by aligning representations across domains. 
The major research line of this category aims to eliminate the domain dependency by minimizing the divergence of feature distributions with different distance metrics, e.g., maximum mean discrepancy~\citep{tzeng2014deep,long2015learning}, an adversarial loss~\citep{ganin2016domain,li2018domain}, Wassertein distance~\citep{zhou2020domain}. Follow-up works applied data augmentation to (1) generate more domains and enhance the consistency of representations during training~\citep{yue2019domain,zhou2020deep,xu2020adversarial,yan2020improve,shu2021open,wang2020heterogeneous,yao2021meta} or (2) generate new domains in an adversarial way to imitate the challenging domains without using training domain information~\citep{zhao2020maximum,qiao2020learning,volpi2018generalizing}. Unlike these latter methods, LISA instead focuses on learning invariant predictors without restricting the internal representations, leading to stronger empirical performance.

\textbf{Learning Invariant Predictors.} Beyond using domain alignment to learning invariant representations, recent work aims to further enhance the correlations between the invariant representations and the labels~\citep{koyama2020out}, leading to invariant predictors. \yao{Representatively, motivated by casual inference, invariant risk minimization (IRM)~\citep{arjovsky2019invariant} and its variants~\citep{guo2021out,khezeli2021invariance,ahuja2021invariance} aim to find a predictor that performs well across all domains through regularizations. Other follow-up works leverage regularizers to penalize the variance of risks across all domains~\citep{krueger2021out}, to align the gradient across domains~\citep{koyama2020out}, to smooth the cross-domain interpolation paths~\citep{chuang2021fair}, or to involve game-theoretic invariant rationalization criterion~\citep{chang2020invariant}. Instead of using regularizers, LISA instead learns domain-invariant predictors via data interpolation.}

\yao{\textbf{Group Robustness.} The last category of methods combating spurious correlations and are particularly suitable for subpopulation shifts. These approaches include directly optimizing the worst-group performance with Distributionally Robust Optimization~\citep{sagawa2019distributionally,zhang2020coping,zhou2021examining}, generating samples around the minority groups~\citep{goel2020model}, and balancing the majority and minority groups via reweighting~\citep{sagawa2020investigation} or regularizing~\citep{cao2019learning,cao2020heteroskedastic}. A few recent approaches in this category target on subpopulation shifts without annotated group labels~\citep{nam2020learning,liu2021just,zhang2021correct,creager2021environment,lee2022diversify}. LISA proposes a more general strategy that is suitable for both domain shifts and subpopulation shifts.}

\section{Conclusion}
To tackle distribution shifts, we propose LISA, a simple and efficient algorithm, to improve the out-of-distribution robustness. LISA aims to eliminate the domain-related spurious correlations among the training set with selective interpolation. We evaluate the effectiveness of LISA on nine datasets under subpopulation shifts and domain shifts settings, demonstrating its promise. Besides, detailed analyses verify that the performance gains caused by LISA result from encouraging learning invariant predictors and representations. Theoretical results further strengthen the superiority of LISA by showing smaller worst-group mis-classification error compared with ERM and vanilla data interpolation. 

While we have made progress in learning invariant predictors with selective augmentation, a limitation of LISA is how to make it compatible with problems in which it is difficult to obtain examples with the same label (e.g., object detection, generative modeling). It would be interesting to explore more general selective augmentation strategies in the future. Additionally, we empirically find that intra-label LISA works without domain information in some domain shift situations. Systematically exploring domain-free intra-label LISA with a theoretical guarantee would be another interesting future direction. 
\section*{Acknowledgement}
We thank Pang Wei Koh for the many insightful discussions. This research was funded in part by JPMorgan Chase \& Co. Any views or opinions expressed herein are
solely those of the authors listed, and may differ from the views and opinions expressed by JPMorgan Chase
\& Co. or its affiliates. This material is not a product of the Research Department of J.P. Morgan Securities
LLC. This material should not be construed as an individual recommendation for any particular client and is
not intended as a recommendation of particular securities, financial instruments or strategies for a particular
client. This material does not constitute a solicitation or offer in any jurisdiction. The research was also supported by Apple and Juniper Networks. The research of Linjun Zhang is partially supported by  NSF DMS-2015378.

\bibliography{ref}
\bibliographystyle{icml2022}
\onecolumn
\appendix
\section{Additional Experiments}
\subsection{Additional Experiments on Subpopulation Shifts}
\subsubsection{Dataset Details}
\label{sec:app_sub_data}

\textbf{Colored MNIST (CMNIST)}: We classify MNIST digits from 2 classes, where classes 0 and 1 indicate original digits (0,1,2,3,4) and (5,6,7,8,9). The color is treated as a spurious attribute. Concretely, in the training set, the proportion between red samples and green samples is 8:2 in class 0, while the proportion is set as 2:8 in class 1. In the validation set, the proportion between green and red samples is 1:1 for all classes. In the test set, the proportion between green and red samples is 1:9 in class 0, while the ratio is 9:1 in class 1. The data sizes of train, validation, and test sets are 30000, 10000, and 20000, respectively. Follow~\cite{arjovsky2019invariant}, we flip labels with probability 0.25.

\textbf{Waterbirds}~\citep{sagawa2019distributionally}: The Waterbirds dataset aims to classify birds as ``waterbird" or ``landbird", where each bird image is spuriously associated with the background ``water" or ``land". 
Waterbirds is a synthetic dataset where each image is composed by pasting a bird image sampled from CUB dataset~\citep{WahCUB_200_2011} to a background drawn from the Places dataset~\cite{zhou2017places}. 
The bird categories in CUB are stratified as land birds or water birds. Specifically, the following bird species are selected to construct the waterbird class: albatross, auklet, cormorant, frigatebird, fulmar, gull, jaeger, kittiwake, pelican, puffin, tern, gadwall, grebe, mallard, merganser, guillemot, or Pacific loon. All other bird species are combined as the landbird class. We define (land background, waterbird) and (water background, landbird) are minority groups. There are 4,795 training samples while only 56 samples are ``waterbirds on land" and 184 samples are ``landbirds on water". The remaining training data include 3,498 samples from ``landbirds on land", and 1,057 samples from ``waterbirds on water".

\textbf{CelebA}~\citep{liu2015faceattributes,sagawa2019distributionally}: 
For the CelebA data~\citep{liu2015faceattributes}, we follow the data preprocess procedure from~\cite{sagawa2019distributionally}.
CelebA defines a image classification task where the input is a face image of celebrities and the classification label is its corresponding hair color --  ``blond” or ``not blond.” The label is spuriously correlated with gender, i.e., male or female. In CelebA, the minority groups are (blond, male) and (not blond, female). The number of samples for each group are 71,629 ``dark hair, female'', 66,874 ``dark hair, male", 22,880 ``blond hair, female", 1,387 ``blond hair, male".

\textbf{CivilComments}~\citep{borkan2019nuanced,koh2021wilds}: 
We use CivilComments from the WILDS benchmark~\citep{koh2021wilds}. 
CivilComments is a text classification task, aiming to predict whether an online comment is toxic or non-toxic. 
The spurious domain identifications are defined as the demographic features, including male, female, LGBTQ, Christian, Muslim, other religion, Black, and White. 
CivilComments contains 450,000 comments collected from online articles. The number of samples for training, validation, and test are 269,038, 45,180, and 133,782, respectively. The readers may kindly refer to Table 17 in~\cite{koh2021wilds} for the detailed group information.

\subsubsection{Training Details}
\label{sec:app_sub_training}
We adopt pre-trained ResNet-50~\citep{he2016deep} and BERT~\citep{sanh2019distilbert} as the model for image data (i.e., CMNIST, Waterbirds, CelebA) and text data (i.e., CivilComments), respectively. In each training iteration, we sample a batch of data per group. 
For intra-label LISA, we randomly apply mixup on sample batches with the same labels but different domains. For intra-domain LISA, we instead apply mixup on sample batches with the same domain but different labels. The interpolation ratio $\lambda$ is sampled from the distribution $\mathrm{Beta}(2,2)$. All hyperparameters are selected via cross-validation and are listed in Table~\ref{tab:hyperameter_sub}.

\subsubsection{Additional Results}
In this section, we have added the full results of subpopulation shifts in Table~\ref{tab:subpopulation_main_full} and Table~\ref{tab:subpopulation_ablation_full}.

\begin{table}[ht]
    \centering
    \small
    \caption{Hyperparameter settings for the subpopulation shifts.}
    \label{tab:hyperameter_sub}
    \begin{tabular}{l|cccc}
    \toprule
        Dataset & CMNIST & Waterbirds & CelebA & CivilComments  \\
        \midrule
        Learning rate &  1e-3 & 1e-3 &  1e-4 & 1e-5\\
        Weight decay & 1e-4 & 1e-4 & 1e-4 & 0 \\
        Scheduler & n/a & n/a & n/a & n/a \\ 
        Batch size & 16 & 16 & 16 & 8 \\
        Type of mixup & mixup & mixup & CutMix & ManifoldMix \\
        Architecture & ResNet50 & ResNet50 & ResNet50 & DistilBert \\
        Optimizer & SGD & SGD & SGD & Adam  \\
        Maximum Epoch & 300 & 300 & 50 & 3 \\
        Strategy sel. prob. $p_{sel}$ & 0.5 & 0.5 & 0.5 & 1.0\\
        \bottomrule
    \end{tabular}
\end{table}

\begin{table*}[h]
\small
\caption{Full results of subpopulation shifts with standard deviation. All the results are performed with three random seed.}
\label{tab:subpopulation_main_full}
\begin{center}
\begin{tabular}{l|cc|cc}
\toprule
\multirow{2}{*}{} & \multicolumn{2}{c|}{CMNIST} & \multicolumn{2}{c}{Waterbirds} \\
& Avg. & Worst  & Avg. & Worst \\\midrule
ERM & 27.8 $\pm$ 1.9\% & 0.0 $\pm$ 0.0\% & 97.0 $\pm$ 0.2\% & 63.7 $\pm$ 1.9\%\\
UW  & 72.2 $\pm$ 1.1\% & 66.0 $\pm$ 0.7\% &  95.1 $\pm$ 0.3\% & 88.0 $\pm$ 1.3\% \\
IRM & 72.1 $\pm$ 1.2\% &  70.3 $\pm$ 0.8\% &  87.5 $\pm$ 0.7\% & 75.6 $\pm$ 3.1\% \\
IB-IRM & 72.2 $\pm$ 1.3\% & 70.7 $\pm$ 1.2\% & 88.5 $\pm$ 0.6\%  & 76.5 $\pm$ 1.2 \% \\
V-REx & 71.7 $\pm$ 1.2\% & 70.2 $\pm$ 0.9\% & 88.0 $\pm$ 1.0\% & 73.6 $\pm$ 0.2\% \\
Coral & 71.8 $\pm$ 1.7\%  & 69.5 $\pm$ 0.9\% & 90.3 $\pm$ 1.1\% & 79.8 $\pm$ 1.8\% \\
GroupDRO & 72.3 $\pm$ 1.2\% & 68.6 $\pm$ 0.8\% & 91.8 $\pm$ 0.3\% & \textbf{90.6 $\pm$ 1.1\%} \\
DomainMix & 51.4 $\pm$ 1.3\% & 48.0 $\pm$ 1.3\% & 76.4 $\pm$ 0.3\% & 53.0 $\pm$ 1.3\% \\
Fish &  46.9 $\pm$ 1.4\% & 35.6 $\pm$ 1.7\% & 85.6 $\pm$ 0.4\% & 64.0 $\pm$ 0.3\%  \\
\midrule
\textbf{LISA} & 74.0 $\pm$ 0.1\% & \textbf{73.3 $\pm$ 0.2\%} & 91.8 $\pm$ 0.3\% & 89.2 $\pm$ 0.6\%  \\\midrule\midrule
& \multicolumn{2}{c|}{CelebA} & \multicolumn{2}{c}{CivilComments} \\
 & Avg. & Worst & Avg. & Worst \\\midrule
ERM  & 94.9 $\pm$ 0.2\% & 47.8 $\pm$ 3.7\% & 92.2 $\pm$ 0.1\% & 56.0 $\pm$ 3.6\% \\
UW & 92.9 $\pm$ 0.2\% & 83.3 $\pm$ 2.8\% & 89.8 $\pm$ 0.5\% & 69.2 $\pm$ 0.9\% \\
IRM &  94.0 $\pm$ 0.4\% & 77.8 $\pm$ 3.9\% &  88.8 $\pm$ 0.7\% & 66.3 $\pm$ 2.1\% \\
IB-IRM & 93.6 $\pm$ 0.3\% & 85.0 $\pm$ 1.8\% &  89.1 $\pm$ 0.3\% & 65.3 $\pm$ 1.5\%  \\ 
V-REx & 92.2 $\pm$ 0.1\% & 86.7 $\pm$ 1.0\%  & 90.2 $\pm$ 0.3\% & 64.9 $\pm$ 1.2\% \\
Coral &  93.8 $\pm$ 0.3\% & 76.9 $\pm$ 3.6\% & 88.7 $\pm$ 0.5\% & 65.6 $\pm$ 1.3\% \\
GroupDRO & 92.1 $\pm$ 0.4\% & 87.2 $\pm$ 1.6\% & 89.9 $\pm$ 0.5\% & 70.0 $\pm$ 2.0\% \\
DomainMix  & 93.4 $\pm$ 0.1\% & 65.6 $\pm$ 1.7\% & 90.9 $\pm$ 0.4\% & 63.6 $\pm$ 2.5\% \\
Fish &  93.1 $\pm$ 0.3\%  &  61.2 $\pm$ 2.5\% & 89.8 $\pm$ 0.4\% & 71.1 $\pm$ 0.4\%  \\\midrule
\textbf{LISA (ours)} & 92.4 $\pm$ 0.4\% & \textbf{89.3 $\pm$ 1.1\%} & 89.2 $\pm$ 0.9\% & \textbf{72.6 $\pm$ 0.1\%}\\
\bottomrule
\end{tabular}
\end{center}
\end{table*}

\begin{table*}[h]
\small
\caption{Full table of the comparison between LISA and other substitute mixup strategies in subpopulation shifts. UW represents upweighting.}
\vspace{-1em}
\label{tab:subpopulation_ablation_full}
\begin{center}
\begin{tabular}{l|cc|cc}
\toprule
\multirow{2}{*}{} & \multicolumn{2}{c|}{CMNIST} & \multicolumn{2}{c}{Waterbirds} \\
& Avg. & Worst  & Avg. & Worst   \\\midrule
ERM & 27.8 $\pm$ 1.9\% & 0.0 $\pm$ 0.0\% & 97.0 $\pm$ 0.2\% & 63.7 $\pm$ 1.9\%\\
Vanilla mixup & 32.6 $\pm$ 3.1\% & 3.1 $\pm$ 2.4\% &  81.0 $\pm$ 0.2\% & 56.2 $\pm$ 0.2\% \\
Vanilla mixup + UW & 72.2 $\pm$ 0.7\% & 71.8 $\pm$ 0.1\% & 92.1 $\pm$ 0.1\% & 85.6 $\pm$ 1.0\% \\
In-group Group & 33.6 $\pm$ 1.9\% & 24.0 $\pm$ 1.1\% & 88.7 $\pm$ 0.3\%  & 68.0 $\pm$ 0.4\% \\
In-group + UW & 72.6 $\pm$ 0.1\% & 71.6 $\pm$ 0.2\% & 91.4 $\pm$ 0.6\% & 87.1 $\pm$ 0.6\% \\
\midrule
\textbf{LISA (ours)} & 74.0 $\pm$ 0.1\% &\textbf{73.3 $\pm$ 0.2\%} & 91.8 $\pm$ 0.3\% & \textbf{89.2 $\pm$ 0.6\%}  \\
\midrule\midrule
& \multicolumn{2}{c|}{CelebA} & \multicolumn{2}{c}{CivilComments} \\
& Avg. & Worst & Avg. & Worst \\\midrule
ERM & 94.9 $\pm$ 0.2\% & 47.8 $\pm$ 3.7\% & 92.2 $\pm$ 0.1\% & 56.0 $\pm$ 3.6\% \\
Vanilla mixup & 95.8 $\pm$ 0.0\% & 46.4 $\pm$ 0.5\% & 90.8 $\pm$ 0.8\% & 67.2 $\pm$ 1.2\% \\
Vanilla mixup + UW & 91.5 $\pm$ 0.2\% & 88.0 $\pm$ 0.3\% & 87.8 $\pm$ 1.2\% & 66.1 $\pm$ 1.4\%  \\
Within Group & 95.2 $\pm$ 0.3\% & 58.3 $\pm$ 0.9\% & 90.8 $\pm$ 0.6\% & 69.2 $\pm$ 0.8\% \\
Within Group + UW & 92.4 $\pm$ 0.4\%  & 87.8 $\pm$ 0.6\% & 84.8 $\pm$ 0.7\% & 69.3 $\pm$ 1.1\% \\
\midrule
\textbf{LISA (ours)} & 92.4 $\pm$ 0.4\% & \textbf{89.3 $\pm$ 1.1\%} & 89.2 $\pm$ 0.9\% & \textbf{72.6 $\pm$ 0.1\%}\\
\bottomrule
\end{tabular}
\end{center}
\end{table*}

\subsection{Additional Experimental Settings on Domain Shifts}
\subsubsection{Dataset Details}
\label{sec:app_domain_data}

In this section, we provide detailed descriptions of datasets used in the experiments of domain shifts and report the data statistics in Table~\ref{tab:domian_data}.
\paragraph{Camelyon17}
We use Camelyon17 from the WILDS benchmark~\citep{koh2021wilds,bandi2018detection}, which provides $450,000$ lymph-node scans sampled from $5$ hospitals. 
Camelyon17 is a medical image classification task where the input $x$ is a $96\times 96$ image and the label $y$ is whether there exists tumor tissue in the image. The domain $d$ denotes the hospital that the patch was taken from. The training dataset is drawn from the first $3$ hospitals, while out-of-distribution validation and out-of-distribution test datasets are sampled from the $4$-th hospital and $5$-th hospital respectively. 

\paragraph{FMoW}
The FMoW dataset is from the WILDS benchmark~\citep{koh2021wilds,christie2018functional} --- a satellite image classification task which includes $62$ classes and $80$ domains ($16$ years x $5$ regions). Concretely, the input $x$ is a $224 \times 224$ RGB satellite image, the label $y$ is one of the $62$ building or land use categories, and the domain $d$ represents the year that the image was taken as well as its corresponding geographical region -- Africa, the Americas, Oceania, Asia, or Europe. 
The train/test/validation splits are based on the time when the images are taken. Specifically, images taken before 2013 are used as the training set. Images taken between 2013 and 2015 are used as the validation set. Images taken after 2015 are used for testing.

\paragraph{RxRx1}
RxRx1~\citep{koh2021wilds,taylor2019rxrx1} from the WILDS benchmark is a cell image classification task. In the dataset, some cells have been genetically perturbed by siRNA. The goal of RxRx1 is to predict which siRNA that the cells have been treated with. Concretely, the input $x$ is an image of cells obtained by fluorescent microscopy, the label $y$ indicates which of the $1,139$ genetic treatments the cells received, and the domain $d$ denotes the experimental batches. Here, $33$ different batches of images are used for training, where each batch contains one sample for each class. The out-of-distribution validation set has images from $4$ experimental batches. The out-of-distribution test set has $14$ experimental batches. The average accuracy on out-of-distribution test set is reported.

\paragraph{Amazon}
Each task in the Amazon benchmark~\citep{koh2021wilds,ni2019justifying} is a multi-class sentiment classification task. The input $x$ is the text of a review, the label $y$ is the corresponding star rating ranging from 1 to 5, and the domain $d$ is the corresponding reviewer. The training set has $245,502$ reviews from $1,252$ reviewers, while the out-of-distribution validation set has $100,050$ reviews from another $1,334$ reviewers. The out-of-distribution test set also has $100,050$ reviews from the rest $1,252$ reviewers. We evaluate the models by the 10th percentile of per-user accuracies in the test set.

\paragraph{MetaShift} 
We use the MetaShift~\citep{metadataset}, which is derived from Visual Genome~\citep{krishnavisualgenome}. 
MetaShift leverages the natural heterogeneity of Visual Genome to provide many distinct data distributions for a given class (e.g. “cats with cars” or “cats in bathroom” for the “cat” class). A key feature of MetaShift is that it provides explicit explanations of the dataset correlation and a distance score to measure the degree of distribution shift between any pair of sets.

We adopt the “Cat vs. Dog” task in MetaShift, where we evaluate the model on the “dog(\emph{shelf})” domain with 306 images, and the “cat(\emph{shelf})” domain with 235 images. The training data for the “Cat” class is the cat(\emph{sofa + bed}), including cat(\emph{sofa}) domain and cat(\emph{bed}) domain. MetaShift provides 4 different sets of training data for the “Dog” class in an increasingly challenging order, i.e., increasing the amount of distribution shift. Specifically, dog(\emph{cabinet + bed}), dog(\emph{bag + box}), dog(\emph{bench + bike}), dog(\emph{boat + surfboard}) are selected for training, where their corresponding distances to dog(\emph{shelf}) are 0.44, 0.71, 1.12, 1.43.

\subsubsection{Training Details}
\label{sec:app_domain_training}
Follow WILDS~\citet{koh2021wilds}, we adopt pre-trained DenseNet121~\citep{huang2017densely} for Camelyon17 and FMoW datasets, ResNet-50~\citep{he2016deep} for RxRx1 and MetaShift datasets, and DistilBert~\citep{sanh2019distilbert} for Amazon datasets.

In each training iteration, we first draw a batch of samples $B_1$ from the training set. With $B_1$, we then select another sample batch $B_2$ with same labels as $B_1$ for data interpolation. The interpolation ratio $\lambda$ is drawn from the distribution $\mathrm{Beta}(2,2)$. We use the same image transformers as~\citet{koh2021wilds}, and all other hyperparameters are selected via cross-validation and are listed in Table~\ref{tab:domain_parameter}.

\begin{table}[ht]
    \centering
    \small
    \caption{Hyperparameter settings for the domain shifts.}
    \begin{tabular}{l|ccccc}
    \toprule
        Dataset & Camelyon17 & FMoW & RxRx1 & Amazon & MetaShift \\
        \midrule
        Learning rate & 1e-4 & 1e-4 & 1e-3 & 2e-6 & 1e-3 \\
        Weight decay & 0 & 0 & 1e-5 & 0 & 1e-4 \\
        Scheduler & n/a & n/a & 
        \shortstack{Cosine Warmup} & n/a & n/a \\
        Batch size & 32 & 32 & 72 & 8 & 16 \\
        Type of mixup & CutMix & CutMix & CutMix & ManifoldMix & CutMix \\
        Architecture & DenseNet121 & DenseNet121 & ResNet50 & DistilBert & ResNet50 \\
        Optimizer & SGD & Adam & Adam & Adam & SGD \\
        Maximum Epoch & 2 & 5 & 90 & 3 & 100\\
        Strategy sel. prob. $p_{sel}$ & 1.0 & 1.0 & 1.0 & 1.0 & 1.0\\
        \bottomrule
    \end{tabular}
    \label{tab:domain_parameter}
\end{table}

\subsection{Strength of Spurious Correlation}
\label{sec:app_spurious_strength}
\yao{In Section~\ref{sec:prelim}, the spurious correlation is defined as the association between the domain $d$ and label $y$, measured by  Cramér's V ~\citep{cramer2016mathematical}. Specifically, let $k_{y,d}$ be the number of samples from domain $d$ with label $y$. The Cramér's V is formulated as 
\begin{equation}
\label{eq:strength_spurious}
    V=\sqrt{\frac{\chi^2}{N\min(|Y-1|, |D-1|)}}=\sqrt{\frac{\sum_{y\in \mathcal{Y}, d\in \mathcal{D}}\frac{(k_{y,d}-\tilde k_{y,d})^2}{\tilde k_{y,d}}}{N\min(|\mathcal{Y}|-1|, |\mathcal{D}|-1|)}},
\end{equation}
where $N$ represents the number of samples in the entire dataset and $\tilde k_{y,d}=\frac{\sum_{y\in \mathcal{Y}} k_{y,d}\sum_{d\in \mathcal{D}}{ k_{y,d}}}{\sum_{y\in \mathcal{Y},d \in \mathcal{D}} k_{y,d}}$. Cramér's V varies from 0 to 1 and higher Cramér's V represents stronger correlation.}

\yao{According to Eq.~\eqref{eq:strength_spurious}, we calculate the strength of spurious correlations on all datasets used in the experiments and report the results in Table~\ref{tab:spurious_strength}. Compared with other datasets, the Cramér's V on Camelyon17, FMoW and RxRx1 are significantly smaller, indicating weaker spurious correlations.}

\begin{table*}[h]
\small
\caption{Analysis of the strength of spurious correlations on datasets with subpopulation shifts or domain shifts.}
\label{tab:spurious_strength}
\begin{center}
\begin{tabular}{cccc|ccccc}
\toprule
\multicolumn{4}{c|}{Subpopulation Shifts} & \multicolumn{5}{c}{Domain Shifts}\\
CMNIST & Waterbirds & CelebA & CivilComments & Camelyon17 & FMoW & RxRx1 & Amazon & MetaShift \\\midrule
0.6000 & 0.8672 & 0.3073 & 0.8723 & 0.0004 & 0.1114 & 0.0067 & 0.3377 & 0.4932 \\
\bottomrule
\end{tabular}
\end{center}
\end{table*}

\subsection{Results on Datasets without Spurious Correlations}
\label{sec:app_no_spurious}
In order to analyze the factors that lead to the performance gains of LISA, we conduct experiments on datasets without spurious correlations. To be more specific, we balance the number of samples for each group under the subpopulation shifts setting. The results of ERM, Vanilla mixup and LISA on CMNIST, Waterbirds and CelebA are reported in Table \ref{tab:app_no_spurious}. The results show that LISA performs similarly compared with ERM when datasets do not have spurious correlations. If there exists any spurious correlation, LISA significantly outperforms ERM. Another interesting finding is that Vanilla mixup outperforms LISA and ERM without spurious correlations, while LISA achieves the best performance with spurious correlations. This finding strengthens our conclusion that the performance gains of LISA are from eliminating spurious correlations rather than simple data augmentation.

\begin{table}[h]
\small
\caption{Results on datasets without spurious correlations. LISA performs similarly to ERM when there are no spurious correlations. However, Vanilla mixup outperforms LISA and ERM when there are no spurious correlations while underperforms LISA on datasets with spurious correlations. The results further strengthen our claim that the performance gains of LISA are not from simple data augmentation.}
\label{tab:app_no_spurious}

\begin{center}
\setlength{\tabcolsep}{1.3mm}{
\begin{tabular}{l|ccc}
    \toprule
        Dataset & CMNIST & Waterbirds & CelebA \\
        \midrule
        ERM & 73.67\% & 88.07\% & 86.11\% \\
        Vanilla mixup & 74.28\% & 88.23\% & 88.89\% \\
        LISA & 73.18\% & 87.05\% & 87.22\% \\
        \bottomrule
    \end{tabular}
}
\end{center}

\end{table}

\subsection{Additional Invariance Analysis}
\subsubsection{Additional Metrics of Invariant Predictor Analysis}
\label{sec:app_additional_predictor_invariance}
In Table~\ref{tab:additional_invariance_predictor}, we report two additional metrics to measure the invariance of predictors -- Risk Variance and Gradient Norm, which is defined as:
\begin{itemize}[leftmargin=*]
    \item \textbf{Risk Variance ($\mathrm{IP}_{var}$)}. Motivated by~\citet{krueger2021out}, we use the variance of test risks across all domains to measure the invariance, which is defined as $\mathrm{IP}_{var}=\mathrm{Var}(\{\mathcal{R}_1(\theta),\ldots, \mathcal{R}_D(\theta)\})$, where $D$ represents the number of test domains and $\mathcal{R}_d(\theta)$ represents the risk of domain $d$.
    \item \textbf{Gradient Norm ($\mathrm{IP}_{norm}$)}. Follow IRMv1~\cite{arjovsky2019invariant}, we use the gradient norm of the classifier to measure the optimality of the dummy classifier at each domain $d$. Assume the classifier is parameterized by $w$, $\mathrm{IP}_{norm}$ is defined as: $\mathrm{IP}_{norm}=\frac{1}{|\mathcal{D}|}\sum_{d\in \mathcal{D}}\| \nabla_{w|w=1.0} \mathcal{R}_d (\theta)\|^2$.
\end{itemize}

\begin{table}[h]
\small
\caption{Additional Invariance Metrics for Invariant Predictor Analysis. We report the results of risk variance ($\mathrm{IP}_{var}$) and gradient norm ($\mathrm{IP}_{norm}$), where smaller values indicate stronger invariance.}
\label{tab:additional_invariance_predictor}
\begin{center}
\setlength{\tabcolsep}{0.9mm}{
\begin{tabular}{l|c|c|c|c|c|c|c|c}
\toprule
& \multicolumn{4}{c|}{$\mathrm{IP}_{var} \downarrow$} & \multicolumn{4}{c}{$\mathrm{IP}_{norm} \downarrow$} \\\cmidrule{2-9}
  & CMNIST & Waterbirds & Camelyon  & MetaShift & CMNIST & Waterbirds  & Camelyon & MetaShift\\\midrule
ERM  & 12.0486 & 0.2456 & 0.0150 & 1.8824  & 1.1162 & 1.5780  & 1.2959 & 1.0914\\
Vanilla mixup & 0.2769 & 0.1465 & 0.0180 & 0.2659  & 1.5347 & 1.8631 & 0.3993 & 0.1985 \\
IRM & 0.0112 & 0.1243 & 0.0201 & 0.8748 & 0.0908 & 0.9798  & 0.5266 & 0.2320 \\
IB-IRM & 0.0072 & 0.2069 & 0.0329 & 0.5680 & 0.6225 & 0.8814 & 0.6890 & 0.1683\\
V-REx & 0.0056 & 0.1257 & 0.0106 & 0.4220 & 0.0290 & 0.8329 & 0.9641 & 0.3680 \\
\midrule
\textbf{LISA (ours)} & \textbf{0.0012} & \textbf{0.0016} & \textbf{9.97e-5} & \textbf{0.2387} & \textbf{0.0039}  & \textbf{0.0538} & \textbf{0.3081} & \textbf{0.1354} \\ 
\bottomrule
\end{tabular}
}
\end{center}
\end{table}

Comparing LISA to other invariant learning methods, the results of $\mathrm{IP}_{var}$ and $\mathrm{IP}_{norm}$ further confirm that LISA does indeed improve predictor invariance.

\subsubsection{Analysis of Learned Invariant Representations}
\label{app:sec_representation_invariance}
\yao{In this section, we use pairwise divergence of representations ($\mathrm{IR}_{kl}$) to measure representation-level invariance. Specifically, assume the representation before classifier of each sample $(x_i, y_i, d)$ is $h_{i,d}$, we compute the KL divergence of the distribution of representations. Similarly, kernel density estimation is also used to estimate the probability density function $P(h_d^y)$ of representations from domain $d$ with label $y$. Formally, $\mathrm{IR}_{kl}$ is defined as $\mathrm{IR}_{kl}=\frac{1}{|\mathcal{Y}||\mathcal{D}|^2}\sum_{y\in \mathcal{Y}}\sum_{d',d\in \mathcal{D}}\mathrm{KL}(P(h_D^y\mid D=d)|P(h_{D}^y\mid D=d'))$. Smaller $\mathrm{IR}_{kl}$ values indicate more invariant representations with respect to the labels. We report the results on CMNIST, Waterbirds, Camelyon17 and MetaShift in Table~\ref{tab:invariance}. Our key observations are: (1) Compared with ERM, LISA learns stronger representation-level invariance. The potential reason is that a stronger 
invariant predictor implicitly includes stronger invariance representation; (2) LISA provides more invariant representations than other regularization-based invariant predictor learning methods, i.e., IRM, IB-IRM, V-REx, showing its capability in learning stronger invariance.}

\begin{table}[h]
\small
\caption{Results of representation-level invariance $\mathrm{IR}_{kl}$ ($\times 10^8$ for CMNIST), where smaller $\mathrm{IR}_{kl}$ value denotes stronger invariance.}
\centering
\label{tab:invariance}
\begin{tabular}{l|c|c|c|c}
\toprule
  & CMNIST & Waterbirds & Camelyon17 & MetaShift \\\midrule
ERM & 1.683 & 3.592 & 8.213 & 0.632 \\
Vanilla mixup & 4.392 & 3.935 & 7.786 & 0.634 \\
IRM  & 1.905 & 2.413 & 8.169 & 0.627  \\
IB-IRM & 3.178 & 3.306 & 8.824 & 0.646 \\
V-REx & 3.169 & 3.414 & 8.838 & 0.617\\
\midrule
\textbf{LISA (ours)} & \textbf{0.421} & \textbf{1.912} & \textbf{7.570} & \textbf{0.585} \\
\bottomrule
\end{tabular}
\end{table}

\begin{figure}[h]
\centering
  \includegraphics[width=0.7\textwidth]{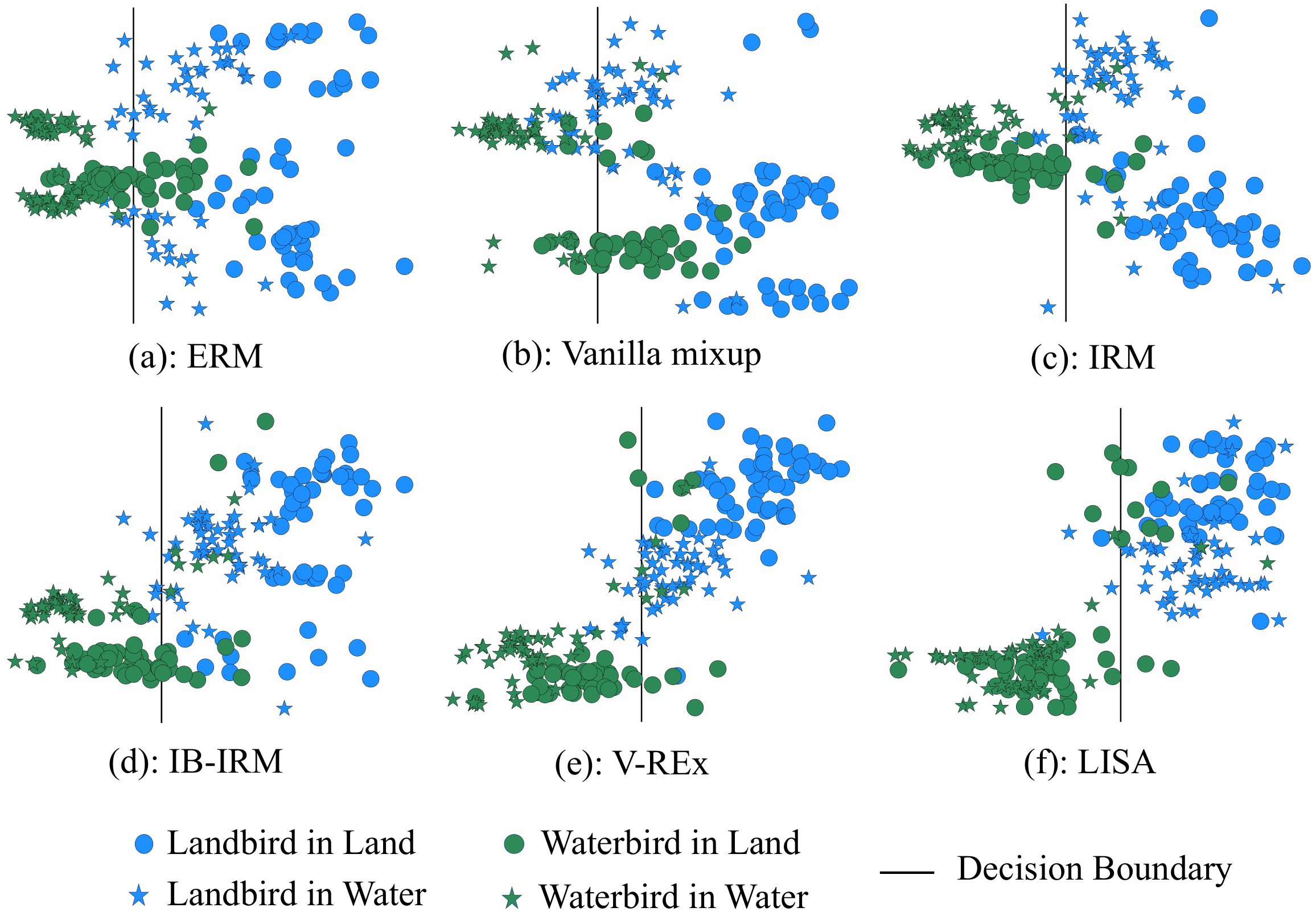}
  \vspace{-0.5em}
  \caption{Visualization of sample representations and decision boundaries on Waterbirds dataset.}\label{fig:boundary}
\end{figure}

\yao{Besides the quantitative analysis, follow Appendix C in~\citet{lee2019meta}, we visualize the hidden representations for all test samples and the decision boundary on Waterbirds and illustrate the results in Figure~\ref{fig:boundary}. Compared with other methods, the representations of samples with the same label that learned by LISA are closer regardless of their domain information, which further demonstrates the promise of LISA in producing invariant representations.}

\subsection{Full Results of WILDS data}
\label{sec:app_domain_full_results}
Follow~\citet{koh2021wilds}, we reported more results on WILDS datasets in Table~\ref{tab:camelyon_full} - Table~\ref{tab:amazon_full}, including validation performance and the results of other metrics. According to these additional results, we could see that LISA outperforms other baseline approaches in all scenarios. Particularly, we here discuss two additional findings: (1) In Camelyon dataset, the test data is much more visually distinctive compared with the validation data, resulting in the large gap ($\sim10\%$) between validation and test performance of ERM (see Table~\ref{tab:camelyon_full}). However, LISA significantly reduces the performance gap between the validation and test sets, showing its promise in improving OOD robustness; (2) In Amazon dataset, though LISA performs worse than ERM in average accuracy, it achieves the best accuracy at the 10th percentile, which is regarded as a more common and important metric to evaluate whether models perform consistently well across all users~\citep{koh2021wilds}.

\begin{table*}[h]
\small
\caption{Full Results of Camelyon17. We report both validation accuracy and test accuracy.}
\label{tab:camelyon_full}
\begin{center}
\begin{tabular}{l|cc}
\toprule
 & Validation Acc. & Test Acc.\\\midrule
ERM & 84.9 $\pm$ 3.1\% & 70.3 $\pm$ 6.4\%  \\
IRM  & \textbf{86.2 $\pm$ 1.4\%} & 64.2 $\pm$ 8.1\% \\
IB-IRM & 80.5 $\pm$ 0.4\% & 68.9 $\pm$ 6.1\% \\
V-REx  & 82.3 $\pm$ 1.3\% & 71.5 $\pm$ 8.3\% \\
Coral & \textbf{
86.2 $\pm$ 1.4\%} & 59.5 $\pm$ 7.7\% \\
GroupDRO & 85.5 $\pm$ 2.2\% & 68.4 $\pm$ 7.3\% \\
DomainMix & 83.5 $\pm$ 1.1\% & 69.7 $\pm$ 5.5\% \\
Fish & 83.9 $\pm$ 1.2\% & 74.7 $\pm$ 7.1\% \\
\midrule
\textbf{LISA (ours)} & 81.8 $\pm$ 1.3\% & \textbf{77.1 $\pm$ 6.5\%} \\
\bottomrule
\end{tabular}
\end{center}
\end{table*}

\begin{table*}[h]
\small
\caption{Full Results of FMoW. Here, we report the average accuracy and the worst-domain accuracy on both validation and test sets.}
\label{tab:fmow_full}
\begin{center}
\begin{tabular}{l|cc|cc}
\toprule
\multirow{2}{*}{}  & \multicolumn{2}{c}{Validation} & \multicolumn{2}{c}{Test}\\
& Avg. Acc. & Worst Acc. & Avg. Acc. & Worst Acc. \\\midrule
ERM &  \textbf{59.5 $\pm$ 0.37\%} & 48.9 $\pm$ 0.62\% & \textbf{53.0 $\pm$ 0.55\%} & 32.3 $\pm$ 1.25\% \\
IRM  & 57.4 $\pm$ 0.37\% & 47.5 $\pm$ 1.57\% & 50.8 $\pm$ 0.13\% & 30.0 $\pm$ 1.37\% \\
IB-IRM & 56.1 $\pm$ 0.48\% & 45.0 $\pm$ 0.62\% & 49.5 $\pm$ 0.49\% & 28.4 $\pm$ 0.90\% \\
V-REx  & 55.3 $\pm$ 1.75\% & 44.7 $\pm$ 1.31\% & 48.0 $\pm$ 0.64\% & 27.2 $\pm$ 0.78\%\\
Coral & 56.9 $\pm$ 0.25\% & 47.1 $\pm$ 0.43\% & 50.5 $\pm$ 0.36\% & 31.7 $\pm$ 1.24\% \\
GroupDRO & 58.8 $\pm$ 0.19\% & 46.5 $\pm$ 0.25\% & 52.1 $\pm$ 0.50\% & 30.8 $\pm$ 0.81\% \\
DomainMix & 58.6 $\pm$ 0.29\% & 48.9 $\pm$ 1.15\% & 51.6 $\pm$ 0.19\% & 34.2 $\pm$ 0.76\% \\
Fish & 57.8 $\pm$ 0.15\% & \textbf{49.5 $\pm$ 2.34\%} & 51.8 $\pm$ 0.32\% & 34.6 $\pm$ 0.18\% \\
\midrule
\textbf{LISA (ours)} & 58.7 $\pm$ 0.92\% & 48.7 $\pm$ 0.74\% & \textbf{52.8 $\pm$ 0.94\%} & \textbf{35.5 $\pm$ 0.65\%} \\
\bottomrule
\end{tabular}
\end{center}
\end{table*}

\begin{table*}[h]
\small
\caption{Full Results of RxRx1. ID: in-distribution; OOD: out-of-distribution}
\label{tab:rr1_full}
\begin{center}
\begin{tabular}{l|ccc}
\toprule
 & Validation Acc. & Test ID Acc. & Test OOD Acc.\\\midrule
ERM & 19.4 $\pm$ 0.2\% & 35.9 $\pm$ 0.4\% & 29.9 $\pm$ 0.4\% \\
IRM  & 5.6 $\pm$ 0.4\% & 9.9 $\pm$ 1.4\% & 8.2 $\pm$ 1.1\% \\
IB-IRM & 4.3 $\pm$ 0.7\% & 7.9 $\pm$ 0.5\% & 6.4 $\pm$ 0.6\% \\
V-REx  & 5.2 $\pm$ 0.6\% & 9.3 $\pm$ 0.9\% & 7.5 $\pm$ 0.8\%\\
Coral & 18.5 $\pm$ 0.4\% & 34.0 $\pm$ 0.3\% & 28.4 $\pm$ 0.3\% \\
GroupDRO & 15.2 $\pm$ 0.1\% & 28.1 $\pm$ 0.3\% & 23.0 $\pm$ 0.3\% \\
DomainMix & 19.3 $\pm$ 0.7\% & 39.8 $\pm$ 0.2\% & 30.8 $\pm$ 0.4\% \\
Fish & 7.5 $\pm$ 0.6\% & 12.7 $\pm$ 1.9\% & 10.1 $\pm$ 1.5\% \\
\midrule
\textbf{LISA (ours)} & \textbf{20.1 $\pm$ 0.4\%} & \textbf{41.2 $\pm$ 1.0\%} & \textbf{31.9 $\pm$ 0.8\%}  \\
\bottomrule
\end{tabular}
\end{center}
\end{table*}

\begin{table*}[h]
\small
\caption{Full Results of Amazon. Both the average accuracy and the 10th Percentile accuracy are reported.}
\label{tab:amazon_full}
\begin{center}
\begin{tabular}{l|cc|cc}
\toprule
\multirow{2}{*}{}  & \multicolumn{2}{c}{Validation} & \multicolumn{2}{c}{Test}\\
& Avg. Acc. & 10-th Per. & Avg. Acc. & 10-th Per. Acc. \\\midrule
ERM &  \textbf{72.7 $\pm$ 0.1\%} & \textbf{55.2 $\pm$ 0.7\%} & 71.9 $\pm$ 0.1\%  & 53.8 $\pm$ 0.8\% \\
IRM  & 71.5 $\pm$ 0.3\% & 54.2 $\pm$ 0.8\% & 70.5 $\pm$ 0.3\% & 52.4 $\pm$ 0.8\% \\
IB-IRM & 72.4 $\pm$ 0.4\% & \textbf{55.1 $\pm$ 0.6\%} & \textbf{72.2 $\pm$ 0.3\%} & 53.8 $\pm$ 0.7\% \\
V-REx  & 72.7 $\pm$ 1.2\% & 53.8 $\pm$ 0.7\% & 71.4 $\pm$ 0.4\% & 53.3 $\pm$ 0.0\% \\
Coral & 72.0 $\pm$ 0.3\% & 54.7 $\pm$ 0.0\% & 70.0 $\pm$ 0.6\% & 52.9 $\pm$ 0.8\% \\
GroupDRO & 70.7 $\pm$ 0.6\% & 54.7 $\pm$ 0.0\% & 70.0 $\pm$ 0.6\% & 53.3 $\pm$ 0.0\% \\
DomainMix & 71.9 $\pm$ 0.2\% & 54.7 $\pm$ 0.0\% & 71.1 $\pm$ 0.1\% & 53.3 $\pm$ 0.0\% \\
Fish & 72.5 $\pm$ 0.0\% & 54.7 $\pm$ 0.0\% & 71.7 $\pm$ 0.1\% & 53.3 $\pm$ 0.0\% \\
\midrule
\textbf{LISA (ours)} & 71.6 $\pm$ 0.4\% & \textbf{55.1 $\pm$ 0.6\%} & 70.8 $\pm$ 0.3\% & \textbf{54.7 $\pm$ 0.0\%} \\
\bottomrule
\end{tabular}
\end{center}
\end{table*}
\renewcommand{\baselinestretch}{1.0}

\def\R{\mathbbm{R}}
\def\P{\mathbbm{P}}
\def\E{\mathbbm{E}}
\def\lam{\lambda}
\def\Sig{\Sigma}
\def\gam{\gamma}
\def\tx{\tilde{x}}
\def\ty{\tilde{y}}
\section{Proofs of Theorem \ref{thm1} and Theorem \ref{thm3}}
\label{sec:app_proof}
\textbf{Outline of the proof}. We will first find the mis-classification errors based on the population version of OLS with different mixup strategies. Next, we will develop the convergence rate of the empirical OLS based on $n$ samples towards its population version. These two steps together give us the empirical mis-classification errors of different methods. We will separately show that the upper bounds in Theorem \ref{thm1} and Theorem \ref{thm3} hold for two selective augmentation strategies of LISA and hence hold for any $p_{sel}\in[0,1]$. Let LL denote intra-label LISA and LD denote intra-domain LISA.

Let $\pi_1=\P(y_i=1)$ and $\pi_0=\P(y_i=0)$ denote the marginal class proportions in the training samples. Let $\pi_R=\P(d_i=R)$ and $\pi_G=\P(d_i=G)$ denote the marginal subpopulation proportions in the training samples. Let $\pi_{G|1}=\P(d_i=G|y_i=1)$ and define $\pi_{G|0}$, $\pi_{R|1}$, and $\pi_{R|0}$ similarly.

We consider the setting where $\alpha:=\pi^{(1,G)}=\pi^{(0,R)}$ is relatively small and $\pi^{(1)}=\pi^{(0)}=\pi^{(G)}=\pi^{(R)}=1/2$.

\subsection{Decomposing the loss function}
Recall that $\Delta=\mu^{(1,G)}-\mu^{(0,G)}=\mu^{(1,R)}-\mu^{(0,R)}$. We further define $\widetilde{\Delta}=\mu^{(1)}-\mu^{(0)}$, $\theta^{(G)}=\mu^{(0,G)}-\E[x_i]$, and $\theta^{(R)}=\mu^{(0,R)}-\E[x_i]$.

For the mixup estimators, we will repeatedly use the fact that $\lam_i$ has a symmetric distribution with support $[0,1]$.

For ERM estimator based on $(X,y)$, where $b_0=\frac{1}{2}-\E[x_i]^Tb$, we have
\begin{align*}
(\mu^{(0,G)})^Tb+b_0&=(\mu^{(0,G)}-\E[x_i])^Tb+\frac{1}{2}\\
&=(\theta^{(G)})^Tb+\E[y_i]\\
(\mu^{(1,G)})^Tb+b_0&=(\mu^{(1,G)}-\E[x_i])^Tb+\frac{1}{2}\\
&=\Delta^Tb+(\theta^{(G)})^Tb+\E[y_i],
\end{align*}

Notice that based on the estimator $b,b_0$, for $d\in\{G,R\}$,
\begin{align}
    \label{err-orginal}
 E^{(1,d)}(b,b_0)=\Phi(\frac{-\Delta^Tb -(\theta^{(d)})^Tb}{\sqrt{b^T\Sig b}})~~\text{and}~~ E^{(0,d)}(b,b_0)=\Phi(\frac{ (\theta^{(d)})^Tb}{\sqrt{b^T\Sig b}}).
\end{align}

In the extreme case where $\pi_{0,R}=\pi_{1,G}=0$, we have 
\[
\widetilde{\Delta}=\mu^{(1,R)}-\mu^{(0,G)},\theta^{(G)}=-\frac{1}{2}\widetilde{\Delta},\theta^{(R)}=\frac{1}{2}\widetilde{\Delta}-\Delta,~\text{and}~\Delta_0:=\mu^{(0,G)}-\mu^{(0,R)}=\Delta-\widetilde{\Delta}.
\]

Hence,
\begin{align}
\label{err0}
    E_0^{(wst)}
    &=\max \{\Phi\left(\frac{(\frac{1}{2}\widetilde{\Delta}-\Delta)^Tb}{\sqrt{b^T\Sig b}}\right),\Phi\left(\frac{-\frac{1}{2}\widetilde{\Delta}^Tb}{\sqrt{b^T\Sig b}}\right)\}.
\end{align}

\subsection{Classification errors of four methods with infinite training samples}
We first provide the limit of the classification errors when $n\rightarrow \infty$.

\subsubsection{Baseline method: ERM}
For the training data, it is easy to show that
\begin{align*}
    var(x)&=\E[var(x|y)]+var(\E[x|y])\\
     &=\Sigma+\E[var(\E[x|y,D]|y)]+var((\mu^{(1)}-\mu^{(0)})y)\\
    &=\Sigma+\E[var(\mu^{(0,R)}-\mu^{(0,G)})\mathbbm{1}(D=R)|y)]+\widetilde{\Delta}^{\otimes 2}\pi^{(1)}\pi^{(0)}\\
    &=\Sigma+\frac{1}{2}(\mu^{(0,R)}-\mu^{(0,G)})^{\otimes 2}(\pi_{R|1}\pi_{G|1}+\pi_{R|0}\pi_{G|0})+\widetilde{\Delta}^{\otimes 2}\pi^{(1)}\pi^{(0)}\\
    cov(x,y)&=cov(\E[x|y],y)\\
    &=cov(\mu^{(0)}+\widetilde{\Delta}y,y)\\
    &=cov(\widetilde{\Delta}y,y)=\widetilde{\Delta}\pi^{(1)}\pi^{(0)}
\end{align*}

For $a_0=\frac{1}{2}(\pi_{R|1}\pi_{G|1}+\pi_{R|0}\pi_{G|0})$ and $\Delta_0=\mu^{(0,G)}-\mu^{(0,R)}$, the ERM has slope and intercept being
\begin{align*}
    b&=var(x)^{-1}cov(x,y)\\
    &\propto (\Sig+ a_0\Delta_0^{\otimes 2})^{-1}\widetilde{\Delta}\\
    &=\Sig^{-1}\widetilde{\Delta}-\Sig^{-1}\Delta_0\cdot \frac{a_0\widetilde{\Delta}^T\Sig^{-1}\Delta_0}{1+a_0\Delta_0^T\Sig^{-1}\Delta_0}\\
    b_0&=\E[y]-\E[x^Tb].
\end{align*}

\subsubsection{Baseline method: Vanilla mixup}
The vanilla mixup does not use the group information. Let $i_1$ be a random draw from $\{1,\dots, n\}$. Let $i_2$ be a random draw from $\{1,\dots, n\}$ independent of $i_1$.
Let
\[
   \tilde{y}_i=\lam_i y_{i_1}+(1-\lam_i)y_{i_2}
\]
and
\[
  \tilde{x}_i=\lam_i x_{i_1}+(1-\lam_i)x_{i_2}.
\]
We can find that
\begin{align*}
    cov(\tx_i,\ty_i)&=cov(\lam_i x_{i_1}+(1-\lam_i)x_{i_2},\lam_i y_{i_1}+(1-\lam_i)y_{i_2})\\
    &=cov(\lam_i x_{i_1},\lam_i y_{i_1})+cov((1-\lam_i)x_{i_2},(1-\lam_i)y_{i_2})\\
    &=(\E[\lam_i^2]+\E[(1-\lam_i)^2])cov(x_i,y_i).\\
    cov(\tx_i)&=(\E[\lam_i^2]+\E[(1-\lam_i)^2])cov(x_i).
\end{align*}
Hence, the population-level slope is the same as the slope in the benchmark method. It is easy to show that the population-level intercept is also the same. Hence,
\begin{align*}
E_{\textup{mix}}^{(wst)}=E_0^{(wst)}.
\end{align*}

\subsection{Intra-label LISA (LISA-L): mixup across domain}

Define
\[
  x_i^{(\lam)}=\lam_i x_{i_1}^{(y_i,G)}+(1-\lam_i)x_{i_2}^{(y_i,R)},
\]
where $i_1$ is a random draw from $\{l: y_l=y_i,D_l=G\}$ and $i_2$ is a random draw from $\{l: y_l=y_i,D_l=R\}$. Then we perform OLS based on $(x_i^{(\lam)},y_i),i=1,\dots,n$.

We can calculate that
\begin{align*}
cov(x_i^{(\lam)},y_i)&=cov(\E[x_i^{(\lam)}|y_i],y_i)=cov(\frac{1}{2}\mu^{(y_i,G)}+\frac{1}{2}\mu^{(y_i,R)},y_i)\\
& =var(y_i)\Delta=\pi^{(1)}\pi^{(0)}\Delta\\
cov(x_i^{(\lam)})&=\E[cov(x_i^{(\lam)}|y_i,\lam_i)]+cov(\E[x_i^{(\lam)}|y_i,\lam_i])\\
&=2\E[\lam_i^2]\Sig+cov(\lam_i(\mu^{(0,G)}-\mu^{(0,R)})+\Delta y_i)\\
&=2\E[\lam_i^2]\Sig+var(\lam_i)(\mu^{(0,G)}-\mu^{(0,R)})^{\otimes 2}+\pi^{(1)}\pi^{(0)}\Delta^{\otimes 2}.
\end{align*}

\subsection{Intra-domain LISA (LISA-D): mixup within each domain}

The interpolated sample can be written as
\begin{align*}
  &(\ty_i,\tx_i)=(\lam_i,\lam_ix_{i_1}^{(1,G)}+(1-\lam_i)x_{i_2}^{(0,G)}) ~\text{if}~d_i=G\\
&(\ty_i,\tx_i)=(\lam_i,\lam_ix_{i_1}^{(1,R)}+(1-\lam_i)x_{i_2}^{(0,R)}) ~\text{if}~d_i=R,
\end{align*}
where $i_1$ is a random draw from $\{l: d_l=d_i,y_i=1\}$ and $i_2$ is a random draw from $\{l: d_l=d_i,y_i=0\}$.

We consider regress $\ty_i$ on $\tx_i$. 
\begin{align*}
cov(\tx_i,\ty_i|d_i=G)&=cov(\E[\tx_i|\ty_i,d_i=G],\ty_i|d_i=G)=var(\ty_i)(\mu^{(1,G)}-\mu^{(0,G)})\\
var(\tx_i|d_i=G)&=\E[var(\tx_i|,\lam_i,D_i=G)|d_i=G]+var(\E[\tx_i|,\lam_i,d_i=G]|D_i=G]\\
&=2\E[\lam_i^2]\Sig+var(\lam_i\mu^{(1,G)}+(1-\lam_i)\mu^{(0,G)}|d_i=G)\\
&=2\E[\lam_i^2]\Sig+var(\ty_i)\Delta^{\otimes 2}.
\end{align*}
We further have
\begin{align*}
    cov(\tx_i,\ty_i)&=\E[cov(\tx_i,\ty_i|d_i)]+cov(\E[\tx_i|d_i],\E[\ty_i|d_i])\\
&=cov(\tx_i^{(G)},\ty_i^{(G)})\pi^{(G)}+cov(\tx_i^{(R)},\ty_i^{(R)})\pi^{(R)}\\
&=var(\ty_i)(\mu^{(1,G)}-\mu^{(0,G)})\pi^{(G)}+var(\ty_i)(\mu^{(1,R)}-\mu^{(0,R)})\pi^{(R)}\\
&=var(\ty_i)\Delta.
\end{align*}
Moreover,
\begin{align*}
    var(\tx_i)&=\E[var(\tx_i|d_i)]+var(\E[\tx_i|d_i])\\
    &=var(\tx_i^{(G)})\pi^{(G)}+var(\tx^{(R)})\pi^{(R)}+(\E[\tx^{(G)}]-\E[\tx^{(R)}])^{\otimes 2}\pi^{(G)}\pi^{(R)}\\
    &=2\E[\lam_i^2]\Sig+var(\lam_i)\Delta^{\otimes 2}+(\mu^{(0,G)}-\mu^{(0,R)})^{\otimes 2}\pi^{(G)}\pi^{(R)}.
\end{align*}

Slope:
\begin{align*}
    b&=var(\tx_i)^{-1}cov(\tx_i,\ty_i)\\
    &\propto (\Sig+a_{\textup{LD}}\Delta_0^{\otimes 2})^{-1}\Delta\\
    &=\Sig^{-1}\Delta-\Sig^{-1}\Delta_0\cdot\frac{a_{\textup{LD}}(\Delta_0)^T\Sig^{-1}\Delta}{1+a_{\textup{LD}}(\Delta_0)^T\Sig^{-1}\Delta_0}\\
    &\propto \Sig^{-1}\tilde{\Delta}+c_{\textup{LD}}\Sig^{-1}\Delta,
\end{align*}
where $a_{\textup{LD}}=\frac{\pi^{(R)}\pi^{(G)}}{2\E[\lam_i^2]}$ and 
 \[
    c_{\textup{LL}}=\frac{1+a_{\textup{LD}}\Delta_0^T\Sig^{-1}\Delta_0-a_{\textup{LD}}\Delta^T\Sig^{-1}\Delta_0}{a_{\textup{LD}}\Delta^T\Sig^{-1}\Delta_0}.
 \]

Moreover, $b_0=\E[\ty_i]-\E[\tx_i]^Tb=\frac{1}{2}-\E[\tx_i]^Tb$.
Notice that
\begin{align*}
    \E[\tx_i]&=\frac{1}{4}(\mu^{(0,G)}+\mu^{(1,G)}+\mu^{(0,R)}+\mu^{(1,R)})\\
    &=\frac{1}{4}(2\mu^{(0,G)}+\Delta+2\mu^{(1,R)}-\Delta)\\
    &=\frac{1}{2}(\mu^{(0,G)}+\mu^{(1,R)})=\E[x_i].
\end{align*}

\underline{\textbf{Method comparison}}.
We only need to compare $E^{(wst)}_{\textup{ERM}}, E^{(wst)}_{LL}, E^{(wst)}_{LD}$.  

For the ERM, $0\leq a_0\leq 2\alpha$ and 
\begin{align*}
b_{\textup{ERM}}&=(1+\frac{a_0\widetilde{\Delta}^T\Sig^{-1}\Delta_0}{1+a_0\Delta_0^T\Sig^{-1}\Delta_0})\Sig^{-1}\widetilde{\Delta}-\frac{a_0\widetilde{\Delta}^T\Sig^{-1}\Delta_0}{1+a_0\Delta_0^T\Sig^{-1}\Delta_0}\Sig^{-1}\Delta\\
&\propto \Sig^{-1}\widetilde{\Delta}-\frac{a_0\widetilde{\Delta}^T\Sig^{-1}\Delta_0}{1+a_0\Delta_0^T\Sig^{-1}\Delta_0+a_0\widetilde{\Delta}^T\Sig^{-1}\Delta_0}\Sig^{-1}\Delta\\
&\propto \Sig^{-1}\widetilde{\Delta}-\frac{a_0\widetilde{\Delta}^T\Sig^{-1}\Delta_0}{1+a_0\Delta^T\Sig^{-1}\Delta_0}\Sig^{-1}\Delta.
\end{align*}
Let $c_0=\frac{a_0\widetilde{\Delta}^T\Sig^{-1}\Delta_0}{1+a_0\Delta^T\Sig^{-1}\Delta_0}$ and $c_1=|c_0|\|\Delta\|_{\Sig}/\|\widetilde{\Delta}\|_{\Sig}$. For simplicity, let $\|v\|_{\Sig}=v^T\Sig^{-1}v$. We first lower bound it via
\begin{align*}
  cor(b_{\textup{ERM}},\widetilde{\Delta})=   \frac{b^T\widetilde{\Delta}}{\|\widetilde{\Delta}\|_{\Sig}\sqrt{b^T\Sig b}}&=\frac{\widetilde{\Delta}^T\Sig^{-1}\widetilde{\Delta}-c_0\Delta^T\Sig^{-1}\widetilde{\Delta}}{\|\widetilde{\Delta}\|_{\Sig}\sqrt{b^T\Sig b}}\\
    &\geq \frac{\widetilde{\Delta}^T\Sig^{-1}\widetilde{\Delta}}{\|\widetilde{\Delta}\|_{\Sig}(\|\widetilde{\Delta}\|_{\Sig}+|c_0|\|\Delta\|_{\Sig})}-\frac{|c_0\Delta^T\Sig^{-1}\widetilde{\Delta}|}{\|\widetilde{\Delta}\|_{\Sig}\sqrt{b^T\Sig b}}\\
    &\geq \frac{1}{1+|c_0|\|\Delta\|_{\Sig}/\|\widetilde{\Delta}\|_{\Sig}}-\frac{c_0\xi\|\Delta\|_{\Sig}}{\|\widetilde{\Delta}\|_{\Sig}-c_0\|\Delta\|_{\Sig}}\\
    &\geq \frac{1-(1+\xi)c_1-c_1^2}{1-c_1^2}=1-C\alpha.
\end{align*}

Similarly, we have
\begin{align*}
   cor(b_{\textup{ERM}},\Delta)=  \frac{b^T\Delta}{\|\Delta\|_{\Sig}\sqrt{b^T\Sig b}}&=\frac{\Delta^T\Sig^{-1}\widetilde{\Delta}-c_0\Delta^T\Sig^{-1}\Delta}{\|\Delta\|_{\Sig}\sqrt{b^T\Sig b}}\\
    &\leq  \frac{\widetilde{\Delta}^T\Sig^{-1}\Delta}{\|\Delta\|_{\Sig}(\|\widetilde{\Delta}\|_{\Sig}\pm c_0\|\Delta\|_{\Sig})}+\frac{|c_0\Delta^T\Sig^{-1}\Delta|}{(\|\widetilde{\Delta}\|_{\Sig}- c_0\|\Delta\|_{\Sig})\|\Delta\|_{\Sig}}\\
    &\leq \frac{1}{1\pm c_0\|\Delta\|_{\Sig}/\|\widetilde{\Delta}\|_{\Sig}}\xi+\frac{c_0\|\Delta\|_{\Sig}/\|\widetilde{\Delta}\|_{\Sig}}{1-c_0\|\Delta\|_{\Sig}/\|\widetilde{\Delta}\|_{\Sig}}\\
    &\leq (\frac{\xi}{1\pm c_1}-\frac{c_1}{1-c_1})\|\Delta\|_{\Sig}.
\end{align*}

Hence,
\begin{align}
\label{err0-lb}
    E_{\textup{ERM}}^{(wst)}\geq \max\left\{\Phi((\frac{1}{2}-C\alpha)\|\widetilde{\Delta}\|_{\Sig}-(\xi+C\alpha)\|\Delta\|_{\Sig}),\Phi((-\frac{1}{2}-C\alpha)\|\widetilde{\Delta}\|_{\Sig})\right\}
\end{align}
 for some constant $C$ depending on the true parameters.

For method LISA-L, using the fact that $\Delta_0=\Delta-\widetilde{\Delta}$, for $a_{\textup{LL}}=var(\lam_i)/(2\E[\lam_i^2)])$,
\begin{align*}
 b_{\textup{LL}}&\propto\Sig^{-1}\Delta+\frac{-a_{\textup{LL}}\Delta^T\Sig^{-1}\Delta_0}{1+a_{\textup{LL}}\Delta_0^T\Sig^{-1}\Delta_0}\Sig^{-1}\widetilde{\Delta}\\
 &\propto \Sig^{-1}\widetilde{\Delta}+c_{\textup{LL}}\Sig^{-1}\Delta
 \end{align*}
 for 
 \[
    c_{\textup{LL}}=\frac{1+a_{\textup{LL}}\Delta_0^T\Sig^{-1}\Delta_0-a_{\textup{LL}}\Delta^T\Sig^{-1}\Delta_0}{a_{\textup{LL}}\Delta^T\Sig^{-1}\Delta_0}=\frac{1-a_{\textup{LL}}\tilde{\Delta}^T\Sig^{-1}\Delta_0}{a_{\textup{LL}}\Delta^T\Sig^{-1}\Delta_0}.
 \]
 Hence,
 \begin{align*}
cor(b_{\textup{LL}},\widetilde{\Delta})&=\frac{ \widetilde{\Delta}^T b_{\textup{LL}}}{\|\widetilde{\Delta}\|_{\Sig}\sqrt{b_{\textup{LL}}^T\Sig b_{\textup{LL}}}} =\frac{\|\widetilde{\Delta}\|_{\Sig}+c_{\textup{LL}}\xi\|\Delta\|_{\Sig}}{\|\widetilde{\Delta}+c_{\textup{LL}}\Delta\|_{\Sig}}\\
 cor(b_{\textup{LL}},\Delta)&=\frac{ b_{\textup{LL}}^T\Delta}{\|\Delta\|_{\Sig}\sqrt{b_{\textup{LL}}^T\Sig b_{\textup{LL}}}}=\frac{\xi\|\widetilde{\Delta}\|_{\Sig}+c_{\textup{LL}}\|\Delta\|_{\Sig}}{\|\widetilde{\Delta}+c_{\textup{LL}}\Delta\|_{\Sig}}.
\end{align*}

To have $E_{\textup{LL}}^{(wst)}< E_{\textup{ERM}}^{(wst)}$, it suffices to require that $(-\frac{1}{2}-C\alpha)\|\widetilde{\Delta}\|_{\Sig}< (\frac{1}{2}-C\alpha)\|\widetilde{\Delta}\|_{\Sig}-(\xi+C\alpha)\|\Delta\|_{\Sig}$ and
\begin{align*}
   &\frac{1}{2} cor(b_{\textup{LL}},\widetilde{\Delta})\|\widetilde{\Delta}\|_{\Sig}-cor(b_{\textup{LL}},\Delta)\|\Delta\|_{\Sig}\leq (\frac{1}{2}-C\alpha)\|\widetilde{\Delta}\|_{\Sig}-(\xi+C\alpha)\|\Delta\|_{\Sig}\\
   &-\frac{1}{2} cor(b_{\textup{LL}},\widetilde{\Delta})\|\widetilde{\Delta}\|_{\Sig}\leq (\frac{1}{2}-C\alpha)\|\widetilde{\Delta}\|_{\Sig}-(\xi+C\alpha)\|\Delta\|_{\Sig}.
\end{align*}

A sufficient condition is
\begin{align*}
    & \xi < (\frac{1}{2}+\frac{1}{2}cor(b_{\textup{LL}},\widetilde{\Delta}))\frac{\|\widetilde{\Delta}\|_{\Sig}}{\|\Delta\|_{\Sig}}-C\alpha,~cor(b_{\textup{LL}},\Delta)\geq \xi+C\alpha,~~cor(b_{\textup{LL}},\widetilde{\Delta})\leq 1-2C\alpha.
\end{align*}
We can find that a further sufficient condition is
\begin{align}
   & \xi < \frac{\|\widetilde{\Delta}\|_{\Sig}}{\|\Delta\|_{\Sig}}-C\alpha, c_{\textup{LL}}> 0, \xi \le \frac{\|\widetilde{\Delta}+c_{\textup{LL}}\Delta\|_{\Sig}-\|\widetilde{\Delta}\|_{\Sig}}{c_{\textup{LL}}\|\Delta\|_{\Sig}}-\eps_1\alpha\label{cond1}\\
    &\|\widetilde{\Delta}+c_{\textup{LL}}\Delta\|_{\Sig}\geq \|\widetilde{\Delta}\|_{\Sig},~\xi\leq \frac{c_{\textup{LL}}\|\Delta\|_{\Sig}}{\|\widetilde{\Delta}+c_{\textup{LL}}\Delta\|_{\Sig}-\|\widetilde{\Delta}\|_{\Sig}}-\eps_1\alpha\label{cond2}\\
   & \xi \leq (\frac{1}{2}+\frac{1}{2}cor(b_{\textup{LL}},\widetilde{\Delta}))\frac{\|\widetilde{\Delta}\|_{\Sig}}{\|\Delta\|_{\Sig}}-C\alpha.\label{cond3}
\end{align}
We first find sufficient conditions for the statements in (\ref{cond1}) and (\ref{cond2}).
Parameterizing $t=c_{\textup{LL}}\|\Delta\|_{\Sig}/\|\widetilde{\Delta}\|_{\Sig}$, we further simplify the condition in (\ref{cond1}) and (\ref{cond2}) as
\begin{align*}
   & \xi < \min\{\frac{\|\widetilde{\Delta}\|_{\Sig}}{\|\Delta\|_{\Sig}},1\}-C\alpha,t(t+2\xi)>0\\
   &\xi\leq \frac{\sqrt{1+t^2+2t\xi}-1}{t}-\eps_1\alpha,~~\xi\leq \frac{1+\sqrt{1+t^2+2t\xi}}{t+2\xi}-\eps_1\alpha.
\end{align*}
We only need to require
\[
  t\geq \max\{0,-2\xi\}~\text{and}~\xi \leq \min\{\frac{\|\widetilde{\Delta}\|_{\Sig}}{\|\Delta\|_{\Sig}},1\}-C\alpha.
\]
Some tedious calculation shows that $t\geq \max\{0,-2\xi\}$ can be guaranteed by
\[
  a_{\textup{LL}}\leq\frac{1}{\|\tilde{\Delta}\|_{\Sig}^2+\|\tilde{\Delta}\|_{\Sig}\|\Delta\|_{\Sig}}~\text{and}~\xi\leq \frac{\|\Delta\|_{\Sig}}{\|\tilde{\Delta}\|_{\Sig}}
\]
It is left to consider the constraint in (\ref{cond3}).
Notice that it holds for any $\xi\leq 0$. When $\xi>0$, we can see
\begin{align*}
cor(b_{\textup{LL}},\widetilde{\Delta})&=\frac{\|\widetilde{\Delta}\|_{\Sig}+\xi c_{\textup{LL}}\|\Delta\|_{\Sig}}{\|\widetilde{\Delta}+c_{\textup{LL}}\Delta\|_{\Sig}}=\frac{1+t\xi}{\sqrt{1+t^2+2t\xi}}\\
&\geq \frac{1+t\xi}{1+t}\geq \xi.
\end{align*}
Hence, it suffices to guarantee that 
\[
   (1-\frac{1}{2}\frac{\|\widetilde{\Delta}\|_{\Sig}}{\|\Delta\|_{\Sig}})\xi<\frac{1}{2}\frac{\|\widetilde{\Delta}\|_{\Sig}}{\|\Delta\|_{\Sig}}-C\alpha.
\]
If $\|\widetilde{\Delta}\|_{\Sig}/\|\Delta\|_{\Sig}\geq 2$, then LHS is negative and it holds. If $1\leq \|\widetilde{\Delta}\|_{\Sig}/\|\Delta\|_{\Sig}< 2$, then the inequality becomes $\xi< 1-C\alpha$. If $\|\widetilde{\Delta}\|_{\Sig}/\|\Delta\|_{\Sig}<1$, then the inequality becomes $\xi\leq \frac{\|\widetilde{\Delta}\|_{\Sig}}{\|\Delta\|_{\Sig}}-C\alpha$.
Because we have required $\xi < \min\{\frac{\|\widetilde{\Delta}\|_{\Sig}}{\|\Delta\|_{\Sig}},1\}-C\alpha$ for some large enough $C$, the constraint (\ref{cond3}) always holds.
To summarize, $E_{\textup{LL}}< E_{\textup{ERM}}$ given that $\xi \leq \min\{\frac{\|\widetilde{\Delta}\|_{\Sig}}{\|\Delta\|_{\Sig}},\frac{\|\Delta\|_{\Sig}}{\|\widetilde{\Delta}\|_{\Sig}}\}-C\alpha$ for some large enough $C$ and $a_{\textup{LL}}\leq 1/(\|\widetilde{\Delta}\|_{\Sig}^2+\|\widetilde{\Delta}\|_{\Sig}\|\Delta\|_{\Sig})$.

For \textbf{method LISA-D}, we can similarly show that $E_{\textup{LD}}\leq E_{\textup{ERM}}$ given that $\xi < \min\{\frac{\|\widetilde{\Delta}\|_{\Sig}}{\|\Delta\|_{\Sig}},\frac{\|\Delta\|_{\Sig}}{\|\widetilde{\Delta}\|_{\Sig}}\}-C\alpha$ for some large enough $C$ and $a_{\textup{LD}}\leq 1/(\|\widetilde{\Delta}\|_{\Sig}^2+\|\widetilde{\Delta}\|_{\Sig}\|\Delta\|_{\Sig})$.

\subsection{Finite sample analysis}
The empirical loss can be written as
\begin{align}
    \label{loss-em}
    &\P(\mathbbm{1}((x_i^{G)})^T\hat{b}+\hat{b}_0>\frac{1}{2})\neq y_i^{(G)})\\
   & =\frac{1}{2}\P((x_i^{G)})^T\hat{b}+\hat{b}_0>\frac{1}{2}|y_i^{(G)}=0)+\frac{1}{2}\P((x_i^{G)})^T\hat{b}+\hat{b}_0<\frac{1}{2}|y_i^{(G)}=1),\nonumber
\end{align}
where
\begin{align*}
    &\P((x_i^{G)})^T\hat{b}+\hat{b}_0>\frac{1}{2}|y_i^{(G)}=0)=\Phi(-\frac{\frac{1}{2}-(\mu^{(0,G)})^T\hat{b}-\hat{b}_0}{\sqrt{\hat{b}^T\Sig \hat{b}}}).\\
    &\P((x_i^{G)})^T\hat{b}+\hat{b}_0<\frac{1}{2}|y_i^{(G)}=1)=\Phi(\frac{\frac{1}{2}-(\mu^{(1,G)})^T\hat{b}-\hat{b}_0}{\sqrt{\hat{b}^T\Sig \hat{b}}}).
\end{align*}

First notice that
\[
  \hat{b}_0=\bar{y}-\bar{x}^T\hat{b}.
\]

We have
\begin{align*}
(\mu^{(0,G)})^T\hat{b}+\hat{b}_0&=(\mu^{(0,G)}-\bar{x})^T\hat{b}+\bar{y}\\
&=(\mu^{(0,G)}-\E[x_i])^T\hat{b}+\frac{1}{2}+\underbrace{\{(\bar{y}-\bar{x}^T\hat{b})-(\E[y_i]-\E[x_i]^T\hat{b})\}}_{R_1}\\
(\mu^{(1,G)})^T\hat{b}+\hat{b}_0&=(\mu^{(1,G)}-\bar{x})^T\hat{b}+\bar{y}\\
&=\Delta^T\hat{b}+(\mu^{(0,G)}-\E[x_i])^T\hat{b}+\frac{1}{2}+R_1.
\end{align*}

Therefore, according to (\ref{loss-em}),
\begin{align*}
   & \frac{1}{2}\Phi(-\frac{\frac{1}{2}-(\mu^{(0,G)})^T\hat{b}-\hat{b}_0}{\sqrt{\hat{b}^T\Sig \hat{b}}})+\frac{1}{2}\Phi(\frac{\frac{1}{2}-(\mu^{(1,G)})^T\hat{b}-\hat{b}_0}{\sqrt{\hat{b}^T\Sig \hat{b}}})\\
    =&\frac{1}{2}\Phi(\frac{(\theta^{(G)})^T\hat{b}+R_1}{\sqrt{\hat{b}^T\Sig \hat{b}}})+\frac{1}{2}\Phi(-\frac{\Delta+(\theta^{(G)})^T\hat{b}+R_1}{\sqrt{\hat{b}^T\Sig \hat{b}}})\\
    =&\frac{1}{2}\Phi(\frac{(\theta^{(G)})^T\hat{b}+R_1}{\sqrt{\hat{b}^T\Sig \hat{b}}})+\frac{1}{2}\Phi(-\frac{(\theta^{(G)})^T\hat{b}+R_1}{\sqrt{\hat{b}^T\Sig \hat{b}}})\\
    &-\left\{\frac{1}{2}\Phi(-\frac{(\theta^{(G)})^T\hat{b}+R_1}{\sqrt{\hat{b}^T\Sig \hat{b}}})-\frac{1}{2}\Phi(-\frac{\Delta+(\Theta^{(G)})^T\hat{b}+R_1}{\sqrt{\hat{b}^T\Sig \hat{b}}})\right\}\\
    =&\frac{1}{2}-\left\{\frac{1}{2}\Phi(-\frac{(\theta^{(G)})^T\hat{b}+R_1}{\sqrt{\hat{b}^T\Sig \hat{b}}})-\frac{1}{2}\Phi(-\frac{\Delta^T\hat{b}+(\theta^{(G)})^T\hat{b}+R_1}{\sqrt{\hat{b}^T\Sig \hat{b}}})\right\}.
\end{align*}

Then the mis-classification error can be written as
\begin{align}
\label{loss2}
    \frac{1}{2}-\frac{1}{2}\underbrace{\left\{\Phi(\frac{(\theta^{(G)})^T\hat{b}+R_1}{\sqrt{\hat{b}^T\Sig \hat{b}}})-\Phi(\frac{(\theta^{(G)})^T\hat{b}-\Delta^T\hat{b}+R_1}{\sqrt{\hat{b}^T\Sig \hat{b}}})\right\}}_{\widehat{L}(\hat{b})}.
\end{align}
Larger the $\widehat{L}(\hat{b})$, smaller the mis-classification error.

We first find that
\begin{align*}
    \widehat{L}(\hat{b})-L(b)\leq C\underbrace{|\frac{(\theta^{(G)})^T\hat{b}+R_1}{\sqrt{\hat{b}^T\Sig \hat{b}}}-\frac{(\theta^{(G)})^Tb}{\sqrt{b^T\Sig b}}|}_{T_1}+C\underbrace{|\frac{(\theta^{(G)})^T\hat{b}-\Delta^T\hat{b}+R_1}{\sqrt{\hat{b}^T\Sig \hat{b}}}-\frac{(\theta^{(G)})^Tb-\Delta^Tb}{\sqrt{b^T\Sig b}}|}_{T_2}.
\end{align*}
In the event that
\[
   \|\Sig^{1/2}(\hat{b}-b)\|_2=o(1) ~\max_{y,d}\|\mu^{(y,d)}\|_2\leq C,~\Sig~\text{is positive definite.}
\]
for the denominator, we have
\begin{align*}
|b^T\Sig b-\hat{b}^T\Sig\hat{b}|&\leq (2\|\Sig^{1/2} b\|_2+\|\Sig^{1/2}(\hat{b}-b)\|_2)\|\Sig^{1/2}(\hat{b}-b)\|_2\\
&\leq 2(1+o(1))\|\Sig^{1/2} b\|_2\|\Sig^{1/2}(\hat{b}-b)\|_2\\
  | \sqrt{\hat{b}^T\Sig\hat{b}}-\sqrt{b^T\Sig b}|&\leq \frac{|\hat{b}^T\Sig\hat{b}-b^T\Sig b|}{\sqrt{\hat{b}^T\Sig\hat{b}}+\sqrt{b^T\Sig b}}\\
  &\leq 2(1+o(1))\|\Sig^{1/2}(\hat{b}-b)\|_2.
\end{align*}

For the numerator, we have
\[
|\frac{1}{2}\widetilde{\Delta}^T\hat{b}+R_1-\frac{1}{2}\widetilde{\Delta}^Tb|\leq |R_1|+\frac{1}{2}\|\Sig^{-1/2}\widetilde{\Delta}\|_2\|\Sig^{1/2}(\hat{b}-b)\|_2.
\]

We arrive at
\begin{align*}
    T_1\leq (1+o(1))\frac{|R_1|+\frac{1}{2}\|\Sig^{-1/2}\widetilde{\Delta}\|_2\|\Sig^{1/2}(\hat{b}-b)\|_2}{\|\Sig^{1/2}b\|_2}+(1+o(1))\frac{|\widetilde{\Delta}^Tb|}{\sqrt{b^T\Sig b}}\frac{\|\Sig^{1/2}(\hat{b}-b)\|_2}{\sqrt{b^T\Sig b}}.
\end{align*}
\begin{align*}
    T_2\leq& (1+o(1))\frac{|R_1|+\frac{1}{2}(\|\Sig^{-1/2}\widetilde{\Delta}\|_2+\|\Sig^{-1/2}\Delta\|_2)\|\Sig^{1/2}(\hat{b}-b)\|_2}{\|\Sig^{1/2}b\|_2}\\
    &+(1+o(1))\frac{|\frac{1}{2}\widetilde{\Delta}^Tb-\Delta^Tb|}{\sqrt{b^T\Sig b}}\frac{\|\hat{b}-b\|_2}{\sqrt{b^T\Sig b}}.
\end{align*}
Moreover $R_1\leq \|\hat{b}-b\|_2+O_P(\frac{1}{\sqrt{n}})$.
To summarize,
\[
 |\widehat{L}(\hat{b})-L(b)|\lesssim (1+o(1))(\|\hat{b}-b\|_2+\frac{1}{\sqrt{n}}).
\]

In the following, we will upper bound $\|\hat{b}-b\|_2$ for each method.
For the \underline{\textbf{ERM method}}, 
\begin{align*}
    \hat{b}=\{(X-\bar{X})^T(X-\bar{X})\}^{-1}(X-\bar{X})^T(y-\bar{y}).
\end{align*}
It is easy to show that
\[
 \|\hat{b}-b\|_2^2=O_P(\frac{p\sum_{i=1}^Nvar(y_i|x_i)}{N^2})=O_P(\frac{p}{N}).
\]

For the \underline{\textbf{vanilla mixup method}}, we first see that
\begin{align*}
  \frac{1}{n}\sum_{i=1}^n\tx_i&=\frac{1}{n}\sum_{i=1}^n(\lam_i x_{i_1}+(1-\lam_i)x_{i_2})=\bar{x}+O_P(n^{-1/2})=\mu+O_P(n^{-1/2})\\
  \frac{1}{n}\sum_{i=1}^n\ty_i&=\pi^{(1)}+O_P(n^{-1/2}).
\end{align*}
Next,
\begin{align*}
    \frac{1}{n}\sum_{i=1}^n\tx_i\ty_i&=  \frac{1}{n}\sum_{i=1}^n\left\{\lam_i^2x_{i_1}y_{i_1}+(1-\lam_i)^2x_{i_2}y_{i_2}+\lam_i(1-\lam_i)x_{i_1}y_{i_2}+\lam_i(1-\lam_i)x_{i_2}y_{i_i}\right\}\\
  \frac{1}{n}\sum_{i=1}^n\tx_i\ty_i- \E[\tx_i\ty_i]  &=\underbrace{\frac{1}{n}\sum_{i=1}^n \tx_i\ty_i-\E[\tx_i\ty_i|X,y]}_{E_1}+\underbrace{\E[\tx_i\ty_i|X,y]-\E[\tx_i\ty_i]}_{E_2}.
\end{align*}
For $E_2$,
\begin{align*}
    E_2&=\frac{2\E[\lam_i^2]}{n}\sum_{i=1}^nx_iy_i-\E[\tx_i\ty_i]=2\E[\lam_i^2]\E[x_iy_i].
\end{align*}
Hence,
\[
  \|E_2\|_2^2=O_P(\frac{p}{n}).
\]

For $E_1$, conditioning on $(X,y)$, $\lam_i^2x_{i_1}y_{i_1}-\frac{\E[\lam_i^2]}{n}\sum_{i=1}^nx_iy_i$ are independent sub-Gaussian vectors. The sub-Gaussian norm of $\frac{1}{N}\sum_{i=1}^n\lam_i^2x_{i_1,j}y_{i_1}-\frac{\E[\lam_i^2]}{n}\sum_{i=1}^nx_{i,j}y_i$ (conditioning on $(X,y)$) can be upper bounded by $c\max_{i\leq N}|x_{i,j}|/\sqrt{n}$. Hence
\begin{align*}
   \P(\|E_1\|_2\geq t|X,y)\leq 2\exp\{-\frac{c_2nt^2}{\max_{j=1}^p\max_{i\leq N}x_{i,j}^2}\}.
\end{align*}
As $x_{i,j}$ are Gaussian distributed, we know that
\[
\P(\sum_{j=1}^p\max_{i\leq n}x_{i,j}^2\geq p\log n)\leq \exp\{-c_3\log n\}.
\]
Hence,
with probability at least $1-\exp(-c_1\log n)$, 
\[
   E_1\leq \frac{Cp\log n}{n}.
\]
To summarize,
\[
   \left\|\frac{1}{n}\sum_{i=1}^n\tx_i\ty_i-(\frac{1}{n}\sum_{i=1}^n\tx_i)(\frac{1}{n}\sum_{i=1}^n\ty_i)-cov(\tx_i,\ty_i)\right\|_2^2=O_P(\frac{p\log n}{n}).
\]
Similarly, we can show that
\[
   \left\|\frac{1}{n}\sum_{i=1}^n\tx_i\tx_i^T-(\frac{1}{n}\sum_{i=1}^n\tx_i)(\frac{1}{n}\sum_{i=1}^n\tx_i)^T-cov(\tx_i)\right\|_2^2=O_P(\frac{p\log n}{n}).
\]
Hence,
\begin{align*}
    \|\hat{b}-b\|_2^2=O_P(\frac{p\log n}{n}).
\end{align*}

For the \underline{\textbf{LISA-L}}, we first see that
\begin{align*}
  \frac{1}{n}\sum_{i=1}^nx^{(\lam)}_i&=\frac{1}{n}\sum_{y_i=1}(\lam_i x_{i_1}^{(1,G)}+(1-\lam_i)x_{i_2}^{(1,R)})+\frac{1}{n}\sum_{y_i=0}(\lam_i x_{i_1}^{(0,G)}+(1-\lam_i)x_{i_2}^{(0,R)})\\
  &=\frac{1}{2}(\bar{x}^{(1,G)}+\bar{x}^{(1,R)})\hat{\pi}_1+\frac{1}{2}(\bar{x}^{(0,G)}+\bar{x}^{(0,R)})\hat{\pi}_0
\end{align*}
We have
\begin{align*}
    \frac{1}{n}(X^{(\lam)})^Ty- \bar{y}\frac{1}{n}\sum_{i=1}^nx^{(\lam)}_i-cov(x_i^{(\lam)},y_i)&=\underbrace{\    \frac{1}{n}(X^{(\lam)})^Ty- \bar{y}\frac{1}{n}\sum_{i=1}^nx^{(\lam)}_i-cov(x_i^{(\lam)},y_i|X,y)}_{E_1}\\&+\underbrace{cov(x_i^{(\lam)},y_i|X,y)-cov(x_i^{(\lam)},y_i)}_{E_2}
\end{align*}
For $E_2$,
\begin{align*}
    E_2&=\frac{\hat{\pi}_1}{2}(\bar{x}^{(1,G)}+\bar{x}^{(1,R)})-\hat{\pi}_1(\frac{1}{2}(\bar{x}^{(1,G)}+\bar{x}^{(1,R)})\hat{\pi}_1+\frac{1}{2}(\bar{x}^{(0,G)}+\bar{x}^{(0,R)})\hat{\pi}_0)-cov(x_i^{(\lam)},y_i)\\
    &=\frac{1}{2}(\bar{x}^{(1,G)}+\bar{x}^{(1,R)}-\bar{x}^{(0,G)}-\bar{x}^{(0,R)})\hat{\pi}_1\hat{\pi}_0-\pi^{(1)}\pi^{(0)}\Delta.
\end{align*}
It is easy to show that
\[
  \|E_2\|_2^2=O_P\left(\frac{p}{\min_{y,e}n^{(y,e)}}\right).
\]

For $E_1$, conditioning on $X$ and $y$, $x_i^{(\lam)}y_i-\E[x_i^{(\lam)}y_i|X,y]$ are independent sub-Gaussian vectors with mean zero. The sub-Gaussian norm of $\frac{1}{n}\sum_{i=1}^nx_{i,j}^{(\lam)}y_i$ (conditioning on $X$ and $y$) can be upper bounded by $c\max_{i\leq n} |x_{i,j}|/\sqrt{N}$.
\begin{align*}
   &\P( \|E_1\|_2\geq t|X,y)=\P\left(\sum_{j=1}^p|\frac{1}{n}\sum_{i=1}^n\{x_{i,j}^{(\lam)}y_i-\E[x_{i,j}^{(\lam)}y_i|X,y]\}|^2\geq t^2|X,y\right)\\
   &\leq 2\exp\left\{-\frac{c_2nt^2}{\sum_{j=1}^p\max_{i\leq n}x_{i,j}^2}\right\}.
\end{align*}
Hence,
\[
  E_1=O_P(\sqrt{\frac{\sum_{j=1}^p\max_{i\leq n}x_{i,j}^2}{n}})=O_P(\frac{p\log n}{n}).
\]
To summarize,
\[
  \|\frac{1}{n}(X^{(\lam)})^Ty-\E[x_i^{(\lam)}y_i]\|_2^2=O_P(\frac{p}{\min_{y,e}n^{(y,e)}}+\frac{p\log n}{n}).
\]

We can use similar analysis to bound
\[
 \|\frac{1}{N}(X^{(\lam)})^TX^{(\lam)}-\E[x_i^{(\lam)}(x_i^{(\lam)})^T]\|_2.
\]
The sub-exponential norm of $\frac{1}{N}\sum_{i=1}^Nx_{i,j}^{(\lam)}x_{i,k}^{(\lam)}$ (conditioning on $X$) can be upper bounded by $\max_{i\leq N}|x_{i,j}||x_{i,k}|/\sqrt{N}$.
We can show that
\begin{align*}
  \|\frac{1}{n}(X^{(\lam)})^TX^{(\lam)}-\E[x_i^{(\lam)}(x_i^{(\lam)})^T]\|_2=O_P(\frac{p}{\min_{y,e}n^{(y,e)}}+\frac{p\log n}{n}).
\end{align*}

For the \underline{\textbf{LISA-D}}, we first see that
\begin{align*}
  \frac{1}{n}\sum_{i=1}^n\tx_i&=\frac{1}{n}\sum_{D_i=G}(\lam_i x_{i_1}^{(1,G)}+(1-\lam_i)x_{i_2}^{(0,G)})+\frac{1}{n}\sum_{D_i=R}(\lam_i x_{i_1}^{(1,R)}+(1-\lam_i)x_{i_2}^{(0,R)})\\
  &=\frac{1}{2}(\bar{x}^{(1,G)}+\bar{x}^{(0,G)})\hat{\pi}_G+\frac{1}{2}(\bar{x}^{(1,R)}+\bar{x}^{(0,R)})\hat{\pi}_R\\
  \bar{\ty}=\frac{1}{2}.
\end{align*}

Next,
\begin{align*}
    \frac{1}{n}\sum_{i=1}^n\tx_i\ty_i&=  \frac{1}{n}\sum_{D_i=G}\left\{\lam_i^2x_{i_1}^{(1,G)}+\lam_i(1-\lam_i)x_{i_2}^{(0,G)}\right\}+\frac{1}{n}\sum_{D_i=R}\left\{\lam_i^2x_{i_1}^{(1,R)}+\lam_i(1-\lam_i)x_{i_2}^{(0,R)}\right\}\\
  \frac{1}{n}\sum_{i=1}^n\tx_i\ty_i -\bar{\tx}\bar{\ty}-cov(\tx,\ty) &=\underbrace{\frac{1}{n}\sum_{i=1}^n \tx_i\ty_i-\bar{\tx}\bar{\ty}-cov(\tx_i,\ty_i|X,y)}_{E_1}+\underbrace{cov(\tx_i,\ty_i|X,y)-cov(\tx_i,\ty_i)}_{E_2}.
\end{align*}
For $E_2$,
\begin{align*}
    E_2&=\hat{\pi}^{(G)}(\E[\lam_i^2](\bar{x}^{(1,G)}-\bar{x}^{(0,G)})+\frac{1}{2}\bar{x}^{(0,G)})+\hat{\pi}^{(R)}(\E[\lam_i^2](\bar{x}^{(1,R)}-\bar{x}^{(0,R)})+\frac{1}{2}\bar{x}^{(0,R)})-\\
   & \frac{1}{4}(\bar{x}^{(1,G)}+\bar{x}^{(0,G)})\hat{\pi}_G-\frac{1}{4}(\bar{x}^{(1,R)}+\bar{x}^{(0,R)})\hat{\pi}_R-var(\lam_i)\Delta\\
    & =\hat{\pi}^{(G)}var(\lam_i)(\bar{x}^{(1,G)}-\bar{x}^{(0,G)})+\hat{\pi}^{(R)}var(\lam_i)(\bar{x}^{(1,R)}-\bar{x}^{(0,R)})-var(\lam_i)\Delta.
\end{align*}
Notice that $E_2$ is a sub-Gaussian vector with sub-Gaussian norm upper bounded by
\[
   \frac{\hat{\pi}_G^2}{n^{(1,G)}}+\frac{\hat{\pi}_G^2}{n^{(0,G)}}+\frac{\hat{\pi}_R^2}{n^{(1,R)}}+\frac{\hat{\pi}_R^2}{n^{(0,R)}}\leq \frac{4}{n}\max_{y,d}\frac{\pi_d}{\pi_{y|d}}.
\]
Using sub-Gaussian concentration, we can show that
\begin{align*}
    E_2=O_P(\sqrt{\frac{p}{n}\max_{y,d}\frac{\pi_d}{\pi_{y|d}}}).
\end{align*}
Notice that $\max_{y,d}\frac{\pi_d}{\pi_{y|d}}\geq 1$.
For $E_1$, conditioning on $X$ and $y$ $\tx_i\ty_i-\E[\tx_i\ty_i|X,y]$ are independent sub-Gaussian vectors with mean zero. The sub-Gaussian norm of $\frac{1}{n}\sum_{i=1}^n\tx_{i,j}\ty_i$ conditioning on $X$ and $y$ can be upper bounded by $c\max_{i,j}|x_{i,j}|$. Similar analysis on $E_1$ leads to
\begin{align*}
     \frac{1}{n}\sum_{i=1}^n\tx_i\ty_i -\bar{\tx}\bar{\ty}-cov(\tx,\ty)=O_P(\sqrt{\frac{p\log n}{n}}+\sqrt{\frac{p}{n}\max_{y,d}\frac{\pi_d}{\pi_{y|d}}}).
\end{align*}

For the sample covariance matrix, we can also show that
\[
   \left\|\frac{1}{n}\sum_{i=1}^n\tx_i\tx_i^T-(\frac{1}{n}\sum_{i=1}^n\tx_i)(\frac{1}{n}\sum_{i=1}^n\tx_i)^T-cov(\tx_i)\right\|_2^2=O_P(\sqrt{\frac{p\log n}{n}}+\sqrt{\frac{p}{n}\max_{y,d}\frac{\pi_d}{\pi_{y|d}}}).
\]

\subsection{A $\xi$-dependent lower bound for $E^{(wst)}_{\textup{ERM}}-E^{(wst)}_{\textup{LL}}$}
Next, we provide a $\xi$-dependent lower bound for $E^{(wst)}_{\textup{ERM}}-E^{(wst)}_{\textup{LL}}$. Based on our previous analysis
\begin{align*}
 E^{(wst)}_{\textup{ERM}}-E^{(wst)}_{\textup{LL}}&\geq c_1\min\left\{(\frac{1}{2}-C\alpha-\frac{1}{2}cor(b_{\textup{LL}},\tilde{\Delta}))\|\tilde{\Delta}\|_{\Sig}+(cor(b_{\textup{LL}},\Delta)-\xi-C\alpha)\|\Delta\|_{\Sig},\right.\\
 &\quad\quad \left.(\frac{1}{2}-C{\alpha}+\frac{1}{2}cor(b_{\textup{LL}},\tilde{\Delta}))\|\tilde{\Delta}\|_{\Sig}-(\xi+C\alpha)\|\Delta\|_{\Sig}\right\},   
\end{align*}
where $c_1$ is a positive constant given by the derivative of $\Phi(\cdot)$. 
Plugging in the expression of $cor(b_{\textup{LL}},\tilde{\Delta})$ and $cor(b_{\textup{LL}},\Delta)$, we have for the first term of $E^{(wst)}_{\textup{ERM}}-E^{(wst)}_{\textup{LL}}$, it is no smaller than
\begin{align*}
&\frac{1}{2}(1-2C\alpha-\frac{1+\xi t}{\sqrt{1+t^2+2t\xi}})\|\tilde{\Delta}\|_{\Sig}+(\frac{\xi+t}{\sqrt{1+t^2+2t\xi}}-\xi-C\alpha)\|\Delta\|_{\Sig}\\
&\geq \frac{1}{2}\frac{t^2}{(1+t)^2}(1-\xi^2)\|\tilde{\Delta}\|_{\Sig}+\frac{t^2}{(1+t)^2}(1-\xi)\|\Delta\|_{\Sig}-C\alpha(\|\Delta\|_{\Sig}+\|\tilde{\Delta}\|_{\Sig}),
\end{align*}
where the last step is due to the current constraint that $t>\max\{0,-2\xi\}$. For the second term, it is no smaller than
\[
  \|\tilde{\Delta}\|_{\Sig}-\xi\|\Delta\|_{\Sig}-C\alpha(\|\tilde{\Delta}_{\Sig}+\|\Delta\|_{\Sig}).
\]
Notice that $t^2/(1+t^2)\geq \min\{\frac{t^2}{4},\frac{1}{4}\}$. We can show that $t\geq \|\tilde{\Delta}\|_{\Sig}/\|\Delta\|_{\Sig}$, then
\begin{align*}
  E^{(wst)}_{\textup{ERM}}-E^{(wst)}_{\textup{LL}}\geq c_3\min\{(\frac{\|\tilde{\Delta}\|_{\Sig}}{\|\Delta\|_{\Sig}} -\xi)\|\Delta\|_{\Sig},(1-\xi)\|\Delta\|_{\Sig},(1-\xi)\|\tilde{\Delta}\|_{\Sig}^2/\|\Delta\|_{\Sig}\}-c_4\alpha(\|\Delta\|_{\Sig}+\|\tilde{\Delta}\|_{\Sig}).
\end{align*}

\subsection{Domain shifts: Proof of Theorem \ref{thm3}}
It still holds that $\widetilde{\Delta}^*=2(\mu^{(0,*)}-\E[x_i^{(\lam)}])=2(\mu^{(0,*)}-\E[\tx_i])$.
It is easy to show that the worst group mis-classification error for this new environment is
\begin{align}
  E^{(wst,*)}_A=\max\left\{\Phi\left(-\frac{\frac{1}{2}(\widetilde{\Delta}^*)^Tb_A}{\sqrt{b_A^T\Sig b_A}}\right),\Phi\left(\frac{\frac{1}{2}(\widetilde{\Delta}^*)^Tb_A-\Delta^Tb_A}{\sqrt{b_A^T\Sig b_A}}\right)\right\},
\end{align}
where $A\in\{\textup{ERM},\textup{mix}, \textup{LL},\textup{LD}\}$.
Notice that
\[
 \widetilde{\Delta}^*=2\mu^{(0,*)}-(\mu^{(0,G)}+\mu^{(1,R)})=\widetilde{\Delta}+\mu^{(0,*)}-\mu^{(0,G)}
\]
We assume $\|\widetilde{\Delta}^*\|_2=\|\widetilde{\Delta}\|_2$.
Let $\xi^*=cor(\Delta,\widetilde{\Delta}^*)$ and $\gamma=cor(\widetilde{\Delta},\widetilde{\Delta}^*)$.
We have
\begin{align*}
    cor(b_{\textup{ERM}},\widetilde{\Delta}^*)&=\frac{\gam\|\widetilde{\Delta}\|_{\Sig}\|\widetilde{\Delta}^*\|_{\Sig}-c_0\xi^*\|\Delta\|_{\Sig}\|\widetilde{\Delta}^*\|_{\Sig}}{\|\widetilde{\Delta}^*\|_{\Sig}\|\widetilde{\Delta}+c_0\Delta\|_{\Sig}}\\
    &= \frac{\gam \|\widetilde{\Delta}\|_{\Sig}}{\|\widetilde{\Delta}\|_{\Sig}\pm \|c_0\Delta\|_{\Sig}}\pm\frac{|c_0\xi^*|\|\Delta\|_{\Sig}}{\|\widetilde{\Delta}\|_{\Sig}\pm \|c_0\Delta\|_{\Sig}}= \gam\pm C\alpha.
\end{align*}

Hence,
\begin{align}
\label{err0-lb}
    E_{\textup{ERM}}^{(wst)}\geq \max\left\{\Phi((\frac{\gam}{2}-C\alpha)\|\widetilde{\Delta}\|_{\Sig}-(\xi-C\alpha)\|\Delta\|_{\Sig}),\Phi((-\frac{\gam}{2}-C\alpha)\|\widetilde{\Delta}\|_{\Sig})\right\}
\end{align}
 for some constant $C$ depending on the true parameters.

 Hence,
 \begin{align*}
cor(b_{\textup{LL}},\widetilde{\Delta}^*)&=\frac{ (\widetilde{\Delta}^*)^T b_{\textup{LL}}}{\|\widetilde{\Delta}^*\|_{\Sig}\sqrt{b_{\textup{LL}}^T\Sig b_{\textup{LL}}}} =\frac{\gam\|\widetilde{\Delta}\|_{\Sig}+c_{\textup{LL}}\xi^*\|\Delta\|_{\Sig}}{\|\widetilde{\Delta}+c_{\textup{LL}}\Delta\|_{\Sig}}.
\end{align*}

To have $E_{\textup{LL}}^{(wst*)}< E_{\textup{ERM}}^{(wst*)}$, it suffices to require that $(-\frac{\gam}{2}-C\alpha)\|\widetilde{\Delta}\|_{\Sig}< (\frac{\gam}{2}-C\alpha)\|\widetilde{\Delta}\|_{\Sig}-(\xi+C\alpha)\|\Delta\|_{\Sig}$ and
\begin{align*}
   &\frac{1}{2} cor(b_{\textup{LL}},\widetilde{\Delta}^*)\|\widetilde{\Delta}\|_{\Sig}-cor(b_{\textup{LL}},\Delta)\|\Delta\|_{\Sig}\leq (\frac{\gam}{2}-C\alpha)\|\widetilde{\Delta}\|_{\Sig}-(\xi+C\alpha)\|\Delta\|_{\Sig}\\
   &-\frac{1}{2} cor(b_{\textup{LL}},\widetilde{\Delta}^*)\|\widetilde{\Delta}\|_{\Sig}\leq (\frac{\gam}{2}-C\alpha)\|\widetilde{\Delta}\|_{\Sig}-(\xi+C\alpha)\|\Delta\|_{\Sig}.
\end{align*}
A sufficient condition is
\begin{align*}
    & \xi < (\frac{\gam}{2}+\frac{1}{2}cor(b_{\textup{LL}},\widetilde{\Delta}^*))\frac{\|\widetilde{\Delta}\|_{\Sig}}{\|\Delta\|_{\Sig}}-C\alpha,~cor(b_{\textup{LL}},\Delta)\geq \xi+C\alpha,~~cor(b_{\textup{LL}},\widetilde{\Delta}^*)\leq \gam-2C\alpha.
\end{align*}
We can find that a further sufficient condition is
\begin{align}
   & \xi < \frac{1+\gam}{2}\frac{\|\widetilde{\Delta}\|_{\Sig}}{\|\Delta\|_{\Sig}}-C\alpha, c_{\textup{LL}}> 0, \xi^*\leq \frac{\gam(\|\widetilde{\Delta}+c_{\textup{LL}}\Delta\|_{\Sig}-\|\widetilde{\Delta}\|_{\Sig})}{c_{\textup{LL}}\|\Delta\|_{\Sig}}-\eps_1\alpha\label{cond1-da}\\
    &\|\widetilde{\Delta}+c_{\textup{LL}}\Delta\|_{\Sig}\geq \|\widetilde{\Delta}\|_{\Sig},~\xi\leq \frac{c_{\textup{LL}}\|\Delta\|_{\Sig}}{\|\widetilde{\Delta}+c_{\textup{LL}}\Delta\|_{\Sig}-\|\widetilde{\Delta}\|_{\Sig}}-\eps_1\alpha\label{cond2-da}\\
   & \xi \leq (\frac{\gam}{2}+\frac{1}{2}cor(b_{\textup{LL}},\widetilde{\Delta}^*))\frac{\|\widetilde{\Delta}\|_{\Sig}}{\|\Delta\|_{\Sig}}-C\alpha.\label{cond3-da}
\end{align}

We first find sufficient conditions for the statements in (\ref{cond1}) and (\ref{cond2}).
Parameterizing $t=c_{\textup{LL}}\|\Delta\|_{\Sig}/\|\widetilde{\Delta}\|_{\Sig}$, we further simplify the condition in (\ref{cond1-da}) and (\ref{cond2-da}) as
\begin{align*}
   & \xi < \frac{1+\gam}{2}\frac{\|\widetilde{\Delta}\|_{\Sig}}{\|\Delta\|_{\Sig}}-C\alpha,t>0,~\xi^*\leq \frac{\gam(\sqrt{1+t^2+2t\xi}-1)}{t}-\eps_1\alpha,\\
   &-\frac{t}{2}\leq \xi\leq t,~~\xi\leq \frac{1+\sqrt{1+t^2+2t\xi}}{t+2\xi}-\eps_1\alpha.
\end{align*}
We only need to require
\[
  t\geq \max\{0,-2\xi\}~\text{and}~\xi < \min\{\frac{1+\gam}{2}\frac{\|\widetilde{\Delta}\|_{\Sig}}{\|\Delta\|_{\Sig}},1\}-C\alpha, \xi^*\leq \gam\xi.
\]
Some tedious calculation shows that $t\geq\max\{0,-2\xi\}$ can be guaranteed by
\[
 a_{\textup{LL}}\leq \frac{1}{\|\tilde{\Delta}\|_{\Sig}^2+\|\widetilde{\Delta}\|_{\Sig}\|\Delta\|_{\Sig}}~\text{and}~\xi\leq \frac{\|\Delta\|_{\Sig}}{\|\tilde{\Delta}\|_{\Sig}}
\]
It is left to consider the constraint in (\ref{cond3-da}).
Notice that it holds for any $\xi\leq 0$. When $\xi>0$, we can see
\begin{align*}
cor(b_{\textup{LL}},\widetilde{\Delta}^*)&=\frac{\gam\|\widetilde{\Delta}\|_{\Sig}+\xi^* c_{\textup{LL}}\|\Delta\|_{\Sig}}{\|\widetilde{\Delta}+c_{\textup{LL}}\Delta\|_{\Sig}}=\frac{\gam+t\xi^*}{\sqrt{1+t^2+2t\xi}}\\
&\geq \frac{\gam+t\xi^*}{1+t}.
\end{align*}
Hence, it suffices to guarantee that 
\[
  \xi^*+\gam\geq \frac{2\|\Delta\|_{\Sig}}{\|\widetilde{\Delta}\|_{\Sig}}\xi+C\alpha.
\]
To summarize, it suffices to require
\[
  a_{\textup{LL}}\leq \frac{1}{\|\tilde{\Delta}\|_{\Sig}^2+\|\widetilde{\Delta}\|_{\Sig}\|\Delta\|_{\Sig}},0\leq \xi^*\leq \gam\xi,\xi < \min\{\frac{\gam}{2}\frac{\|\widetilde{\Delta}\|_{\Sig}}{\|\Delta\|_{\Sig}},\frac{\|\Delta\|_{\Sig}}{\|\tilde{\Delta}\|_{\Sig}}\}-C\alpha. 
\]

For LISA-D, we can similarly show that $E^{(wst*)}_{\textup{LD}}< E^{(wst*)}_{\textup{ERM}}$ given that
\[
 a_{\textup{LD}}\leq \frac{1}{\|\tilde{\Delta}\|_{\Sig}^2+\|\widetilde{\Delta}\|_{\Sig}\|\Delta\|_{\Sig}},0\leq \xi^*\leq \gam\xi,\xi < \min\{\frac{\gam}{2}\frac{\|\widetilde{\Delta}\|_{\Sig}}{\|\Delta\|_{\Sig}},\frac{\|\Delta\|_{\Sig}}{\|\tilde{\Delta}\|_{\Sig}}\}-C\alpha. 
\]

\end{document}